\documentclass[times,twocolumn,final]{elsarticle}
\usepackage{medima}
\usepackage{framed,multirow}

\usepackage{amssymb}
\usepackage{latexsym}
\usepackage{algorithm}
\usepackage{algpseudocode}
\usepackage{mathtools}
\usepackage{amsmath}
\usepackage{multirow}
\usepackage{booktabs}
\usepackage{wrapfig}
\usepackage{multicol}
\usepackage{subcaption}
\usepackage{url}
\usepackage{xcolor}

\usepackage{hyperref}

\definecolor{newcolor}{rgb}{.8,.349,.1}

\journal{Medical Image Analysis}

\begin{document}

\verso{Xiaoxiao Li \textit{et~al.}}

\begin{frontmatter}

\title{Multi-site fMRI Analysis Using Privacy-preserving Federated Learning and Domain Adaptation: ABIDE Results}%
%\tnotetext[tnote1]{This is an example for title footnote coding.}

\author[1]{Xiaoxiao \snm{Li}\corref{cor1}}
\cortext[cor1]{Corresponding author: 
  Email: xiaoxiao.li@aya.yale.edu }
\author[2]{Yufeng \snm{Gu}\fnref{fn1,fn2}}
\fntext[fn1]{This work was done at Yale University.}
% %% Third author's email
% \ead{author3@author.com}
\author[1,3]{Nicha \snm{Dvornek}\fnref{fn2}}
\fntext[fn2]{YG and ND had equal contribution.}
\author[1,3,5]{Lawrence H. \snm{Staib}}
\author[4]{Pamela \snm{Ventola}}
\author[1,3,5,6]{James S. \snm{Duncan} \corref{cor2}}
\cortext[cor2]{Corresponding author: 
  Email: james.duncan@yale.edu }

\address[1]{Biomedical Engineering, Yale University, New Haven, CT, 06511, USA}
\address[2]{College of Information Science \& Electronic Engineering, Zhejiang University, Hangzhou, 310058, China}
\address[3]{Radiology \& Biomedical Imaging, Yale School of Medicine, New Haven, CT, 06511, USA}
\address[4]{Child Study Center, Yale School of Medicine, New Have, CT, 06511, USA}
\address[5]{Electrical Engineering, Yale University, New Haven, CT, 06511, USA}
\address[6]{Statistics \& Data Science, Yale University New Haven, CT, 06511, USA}
\received{}
\finalform{}
\accepted{}
\availableonline{}
\communicated{}

\begin{abstract}
%%%
Deep learning models have shown their advantage in many different tasks, including neuroimage analysis. However, to effectively train a high-quality deep learning model, the aggregation of a significant amount of patient information is required. The time and cost for acquisition and annotation in assembling, for example, large fMRI datasets make it difficult to acquire large numbers at a single site. %Cooperation from different institutions is desired. 
However, due to the 
need to protect the privacy of patient data,
it is hard to assemble a central database from multiple institutions. Federated learning allows for population-level models to be trained without centralizing entities' data by transmitting the global model to local entities, training the model locally, and then averaging the gradients or weights in the global model. However, some studies suggest that private information can be recovered from the model gradients or weights. In this work, we address the problem of multi-site fMRI classification with a privacy-preserving strategy. To solve the problem, we propose a federated learning approach, where a decentralized iterative optimization algorithm is implemented and shared local model weights are altered by a randomization mechanism. Considering the systemic differences of fMRI distributions from different sites, we further propose two domain adaptation methods in this federated learning formulation. We investigate various practical aspects of federated model optimization and compare federated learning with alternative training strategies. Overall, our results demonstrate that it is promising to utilize multi-site data without data sharing to boost neuroimage analysis performance and find reliable disease-related biomarkers. Our proposed pipeline can be generalized to other privacy-sensitive medical data analysis problems. Our code is publicly
available at: \hyperlink{https://github.com/xxlya/Fed_ABIDE}{https://github.com/xxlya/Fed\_ABIDE/}.
%%%%
\end{abstract}

\begin{keyword}
%% MSC codes here, in the form: \MSC code \sep code
%% or \MSC[2008] code \sep code (2000 is the default)
% \MSC 41A05\sep 41A10\sep 65D05\sep 65D17
%% Keywords
\KWD Federated Learning  \sep Domain Adaptation \sep Data Sharing \sep Privacy \sep rs-fMRI
\end{keyword}

\end{frontmatter}

%\linenumbers

%% main text
\section{Introduction}
\label{sec:intro}

Data has “non-rivalrous” value, a term from the economics literature \citep{weimer2017policy}, meaning that it can be utilized by multiple parties at a time to create additional data products or services. Pooling data together will have synergistic effects. For example, for developing a deep neural network for image recognition tasks, a vast training set is needed that captures the complexity of the problem (in some cases as many as ten thousand images). However, similar data at scale tend not to be available in healthcare, resulting in a lack of generalizability and accuracy for models and concerns regarding the reproducibility of results. Sharing large amounts of medical data is essential for precision medicine, with one important example being functional MRI (fMRI) data related to certain neurological diseases or disorders. The time and cost for acquisition and annotation in gathering large fMRI datasets make it difficult to recruit large numbers at a single site. Deep learning models have shown their advantage in fMRI analysis \citep{suk2016state,shen2017deep}. Without assembling data from a number of different locations, the typically limited amount of data available from a single site becomes an obstacle to building an accurate deep learning model for neuroimage analysis.

However, there are many concerns regarding medical data sharing. For example, patients might be concerned about sharing their medical data, due to the risk that it will be shared with employers or used for future health insurance decision-making if their data are stored and accessed by multiple users, even when deidentified \citep{roski2014creating}.  There are questions about whether deidentified data are truly anonymous. From a legal point of view, data sharing is regulated by different federal and state laws. The power of regulation might vary due to the content of the data, its identifiability, and the context of its use \citep{rosenbaum2005assessing}. Many governmental agencies have their own privacy and data-sharing policies \citep{policy2003cdc}. In addition, health systems are concerned that competitors will be able to use their data when they compete for customers. Providers worry that if their health statistics are publicly available, they will lose patients or be sanctioned if they cannot assess their performance \citep{heitmueller2014developing}.

To tackle the data-sharing problem, \textit{Federated learning} \citep{li2019federated} was introduced to protect privacy by using training data distributed among multiple parties. Instead of transferring data directly to a centralized data warehouse for building machine learning models, in a federated learning setup, each party retains its data and performs decentralized computing. Hence, federated learning addresses privacy concerns and encourages multi-institution collaboration.

\begin{wrapfigure}{r}{0.25\textwidth}
  \centering
    \includegraphics[width=0.25\textwidth]{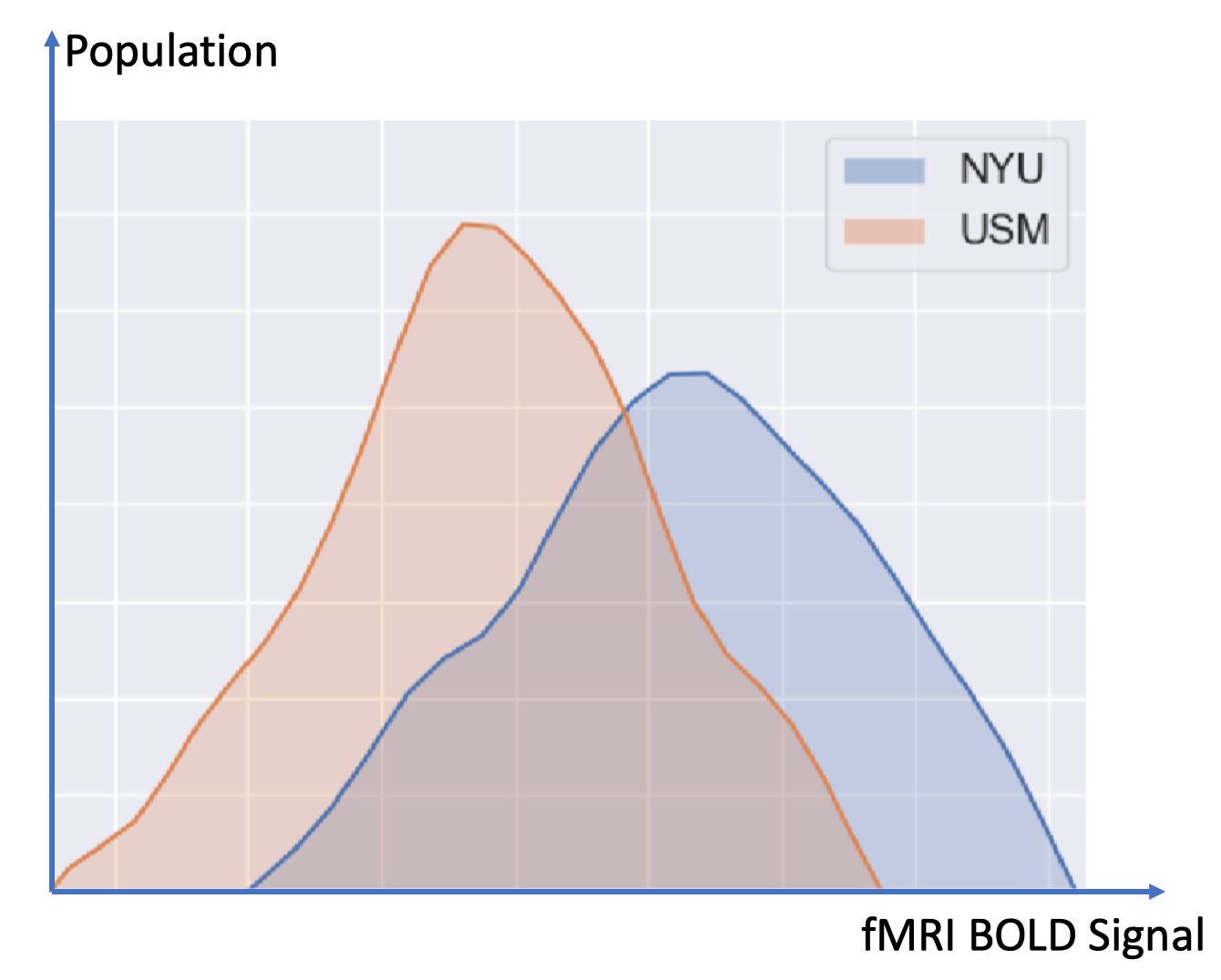}
  \caption{fMRI distribution of different sites}
   \label{fig:dist}
\end{wrapfigure}
Another problem existing in utilizing data from different parties is domain shift. Diverse domains of data are common because institutions can have very different methods of data generation and collection.
The scanners used in different institutions may be from different manufacturers, may be calibrated differently and may have different acquisition protocols specified.
For example, in data from the Autism Brain Imaging Data Exchange (ABIDE I) \citep{di2014autism}, the University of Utah School of Medicine (USM) site used a 3T Siemens TrioTim MR scanner, the New York University (NYU) site used a 3T Siemens Allegra MR scanner, while the University of Michigan (UM) site used a 3T GE Signa MR scanner. Also, the instructions given to each subject were different at different sites. The USM site told participants to "Keep your eyes open and remain awake, letting thoughts pass through your mind without focusing on any particular mental activity" 
while participants at the UM site looked at a fixation cross in the middle of the screen and participants at the NYU site were asked to look at a white cross-hair against a black background that was projected on a screen but some participants' eyes were closed during scanning. Figure \ref{fig:dist} shows the heterogeneous fMRI data distribution of NYU and USM sites, although both of the sites used the scanners from the same manufacturer. One of the challenges of imaging studies of brain disorders is to detect robust findings across sites.Recent studies \citep{yao2019heterogeneous,wang2018deep} have shown promising results of utilizing domain adaptation techniques to assist heterogeneous data analysis, including the applications in medical image analysis areas \citep{chen2019synergistic,yang2019unsupervised}. Therefore, federated learning, together with domain adaptation methods, has the potential to extract reliable, robust neural patterns from brain imaging data of patients having different psychiatric disorders. \\

\textbf{Our contributions} are summarized as follows:
\begin{enumerate}
    \item We formulate a new privacy-preserving pipeline for multi-site fMRI analysis and investigate various practical aspects of the federated model's communication frequency and privacy-preserving mechanisms.
    \item To the best of our knowledge, we investigate domain adaptation in federated learning for medical image analysis for the first time. Domain shift due to heterogeneous data distribution is a challenging issue when utilizing medical images from different institutions. 
%     \item We propose new evaluation metrics to examine explanation performance.
    \item We propose to evaluate performance based on the biomarkers detected by the model, in addition to direct assessment of accuracy metrics.
\end{enumerate}

\textbf{Paper structure:} 
In Section \ref{sec:relate}, we summarize related work about federated learning and unsupervised domain adaptation, the two techniques we focus on in this paper. In Section \ref{sec:method}, we introduce the methods used for our study. Specifically, in Section \ref{sec:fed}, we propose the privacy-preserving federated learning setup for multi-site fMRI analysis; in Section \ref{sec:adaptation}, we propose two domain adaptation methods to boost federated learning performance; and in Section \ref{sec:evalinterp}, we propose the biomarker detection and evaluation methods. The experiments, results, and evaluation methods are presented in Section \ref{sec:experiment}. We conclude the paper in Section \ref{sec:conclusion}.

\section{Related Work} \label{sec:relate}
\subsection{Federated Learning}
Generally, federated learning can be achieved by two approaches: 1) each party training the model using private data and where only model parameters being transferred and 2) using encryption techniques to allow safe communications between different parties \citep{yang2019federated}. In this way, the details of the data are not disclosed in between each party. In this paper, we focus on the first approach, which has been studied in \citep{dean2012large, shokri2015privacy,mcmahan2016}.

Obtaining sufficient data is a major challenge in the field of medical imaging. Apart from data collection, labeling medical image data that require expert knowledge can be addressed by the collaboration between institutions. However, there are lots of potential legal and technical issues when sharing medical data to a centralized location, especially among international institutions. In the medical imaging field, multi-institutional deep learning without sharing patient data was firstly investigated in  \citep{sheller2018multi}. Later, another work \citep{li2019privacy} empirically studied privacy-preserving issues using a sparse vector technique and investigated model weights sharing schemes for imbalanced data. We note that the randomization mechanism for privacy protection and domain adaptation issues have not been studied in federated learning for medical images. We address these two issues in our study.
\subsection{Domain Adaptation}
 Domain Adaptation aims to transfer the knowledge learned from a source domain to a target domain. Then, a model trained over a data set from a source domain is further refined to adapt to a data set from a different target domain. Unsupervised domain adaptation methods have been extensively studied \citep{gholami2018unsupervised, zhao2019multi, hoffman2018algorithms,long2015learning,ganin2014unsupervised,tzeng2017adversarial,zhu2017unpaired,long2018conditional}. However, these efforts cannot meet the requirements of federated settings: data are stored locally and cannot be shared, which hinders adaptive approaches in mainstream domains because they require access to source and target data \citep{tzeng2014deep,long2017deep,ghifary2016deep,sun2016deep,ganin2014unsupervised,tzeng2017adversarial}. Federated domain adaptation has been recently proposed \citep{peng2019federated,peterson2019private}. In our study, we investigate adopting those two federated domain adaptation methods in our multi-source and multi-target federated learning domain adaptation problem.  

\section{Methods}\label{sec:method}
\subsection{Basic privacy-preserving federated learning setup} \label{sec:fed}
In this section, we formulate multi-site fMRI analysis without data sharing in a federated learning framework. Then we introduce the randomized mechanism for privacy protection. Finally, we show the details of training such a privacy-preserving federated learning network step by step.
\subsubsection{Problem definition}
\label{define}
\begin{figure*}[t]
    \centering
    \includegraphics[width=0.8\textwidth]{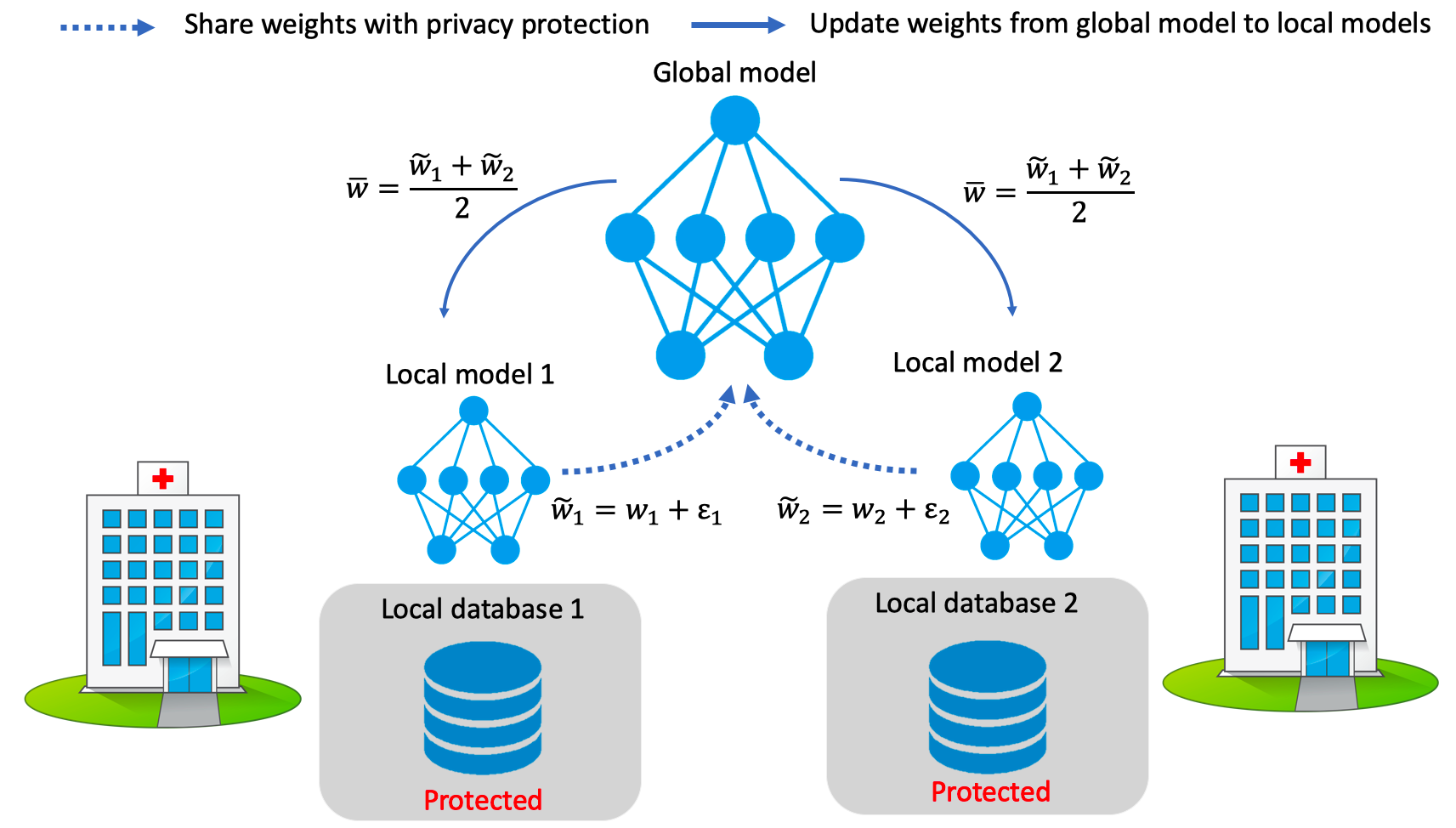}
    \caption{The simplified example of privacy-preserving federated learning strategy for fMRI analysis.}
    \label{fig:framework}
\end{figure*}
Let matrix $\mathcal{D}_i$ denote the data held by the data owner site $i$.  Define $N$ sites $\{\mathcal{F}_1,\dots.\mathcal{F}_N\}$, all of whom wish to train a deep learning model by consolidating their respective data $\{\mathcal{D}_1,\dots.\mathcal{D}_N\}$. For medical imaging problems, usually, the data size at each site is limited to train an accurate deep learning model. A conventional method is to put all data together and use $\mathcal{D} =\mathcal{D}_1 \cup \dots \cup \mathcal{D}_N$ to train a model $\mathcal{M}_{MIX}$.  At the same time, some data sets may also contain label data. We denote the feature space as $\mathcal{X}$, the label space as $\mathcal{Y}$ and we use $\mathcal{I}$ to denote the sample ID space. The feature space  $\mathcal{X}$ , label $\mathcal{Y}$ and sample IDs $\mathcal{I}$ constitute the complete training dataset $(\mathcal{I},\mathcal{X},\mathcal{Y})$. In our multi-site fMRI classification scenario: $\mathcal{D}_i$ is fMRI data, $\mathcal{F}_i$ is the institution owning private fMRI data; $\mathcal{X}$ is the extracted fMRI feature and label $\mathcal{Y}$  can be the diagnosis or phenotype we want to predict. In this setting, data sets share the same feature space but are different in samples. For example, different sites have different subjects. However, the features are all fMRI signals extracted from the same preprocessing pipeline. Therefore, we can summarize the data distribution as:
\begin{equation} \label{eq:problem}
    \mathcal{X}_i = \mathcal{X}_j,\quad \mathcal{Y}_i = \mathcal{Y}_j,\quad\mathcal{I}_i \neq \mathcal{I}_j, \quad \forall \mathcal{D}_i, \mathcal{D}_j, \quad i \neq j,
\end{equation}
which belongs to the horizontal federated learning category where different data sets have large overlap on features while they have small overlap on samples \citep{yang2019federated}. 

In this scenario, due to regulation and other issues, each medical institution will  not share data with the other parties.  A federated learning system is a learning process where the data owners collaboratively train a model $\mathcal{M}_{FED}$, in which any data owner $\mathcal{F}_i$ does not expose its data  $\mathcal{D}_i$ to others.   In our problem setting, assume there is a central server for computing (not for data storage). All the different medical institutions (sites) use the same deep learning architecture for the same task. Each institution trains the deep learning model in-house and updates the model weight information to a central server at a particular frequency during training. The shared weights are blurred by additive random noise $\varepsilon$ to protect data from inverse interpretation leakage. Once the central server receives all the weights, it summarizes them and updates the new weights to each institution. The simplified pipeline is depicted in Figure \ref{fig:framework}.  %Each row of the matrix represents a sample, and each column represents a feature.

\subsubsection{Privacy-preserving decentralized training}
The simplified federated learning framework is depicted in Figure \ref{fig:framework}, which contains two key steps in decentralized optimization: 1) local update, and 2) communicating to a global server. The detailed training procedure is presented in Algorithm \ref{ag1}. The objective function in Algorithm \ref{ag1} for training data in any site $n$ is cross-entropy loss:
\begin{equation} \label{eq:celoss}
    \mathcal{L}_{ce}^{n}=-\sum_{n_i}[y_{n_i}\log(p_{n_i})+(1-y_{n_i})\log(1-p_{n_i})]
\end{equation}
where $y_{n_i}$ is the label of $i$th subject in the training label set $Y_n = \{y_{n_1},\dots,y_{n_{|Y_n|}}\}$ and $p_{n_i}$ is the corresponding model output, which estimates the probability of that label, given an input. All the training inputs and training labels are sampled from feature space  $\mathcal{X}_n$ and label space $\mathcal{Y}_n$.
\begin{algorithm}[htpb]
    		\caption{Privacy-preserving federated learning for multi-site fMRI analysis}\label{ag1}
    		\hspace*{\algorithmicindent} \textbf{Input:} {1. $\mathbf{X}=\{X_1,\dots,X_N\}$, fMRI data from $N$ institutions/sites;  2. $\mathbf{f}_{w}=\{f_{w_1},\dots,f_{w_N}\}$, local models within $N$ sites, where $w_i$ is local model weights;  3. $\mathbf{Y} =\{Y_1,\dots,Y_N\}$, fMRI labels; 4. $M(\cdot)$, noise generator that is used for privacy-preservation (explained in the following section); 5. $K$, number of optimization iterations; 6. $\tau$, global model updating pace, which means the global model and the private models communicate per $\tau$ steps in each optimization iteration; %7. $f_{\bar{w}}$, global model
    		7. $\{opt_1(\cdot),\dots,opt_N(\cdot)\}$, optimizer returning updated model weights w.r.t. objective function $\mathcal{L}$.
    			%\hspace*{\algorithmicindent} \textbf{Output}
    			\begin{algorithmic}[1]
    				\State {$\{{w_1^{(0)}},\dots,{w_N^{(0)}}\} \gets \text{randomize parameters} $}
    				\Comment{initialize local model}
    				\For{$k = 1$ to $K$}
    				\State{$t \gets 0$}
    				\Comment{initialize pace counter}
    				\For{$n = 1$ to $N$}
    				\State{$w_n^{(k)} \gets opt_n(\mathcal{L}(f_{w_n^{(k-1)}}(X_n,Y_n))$}
    				\EndFor
    				\State{$t \gets t+1$}
    				\Comment{models communicate}
    				\If{$t \% \tau = 0$}
    				\State{$\bar{w}^{(k)} \gets \frac{1}{N}\sum_n (w_n^{(k)}+ M(w_n^{(k)})) $}
    				\Comment{update global model per $\tau$ steps}
    				\For{$n = 1$ to $N$}
    				\State{$w_n^{(k)} \gets \bar{w}^{(k)}$}
    				\Comment{deploy weights to local model }
    				\EndFor
    				\EndIf
    				\EndFor	
    				%\State{$\mathcal{M}_{ij} \gets \mathcal{M}_{ij}\mathcal{G}_{ij}$}
    				% \State{$\mathcal{M}_{ij} \gets \frac{\textit{exp}(\mathcal{M}_{ij}A_{ij})}{\sum_{j}\textit{exp}(\mathcal{M}_{ij}A_{ij})}$}
    				% \Comment{calculate probability mask}				
    			\end{algorithmic}
    			\hspace*{\algorithmicindent} \textbf{Return: global model} $f_{\bar{w}^{(K)}}$
    		}
        \end{algorithm}

\subsubsection{Randomized mechanism for privacy protection}
Differential privacy \citep{dwork2014algorithmic,dwork2006calibrating}  is a popular approach to privacy-preserving machine learning \citep{shokri2015privacy} and establishes a strong standard for privacy guarantees for aggregated database-based algorithms. Informally, differential privacy aims to provide
a bound, $\epsilon$, that the attacker could learn virtually nothing more about an individual than they would learn if it were absent from the dataset as the individual's sensitive information is almost irrelevant in the outputs of the model.  The bound $\epsilon$ represents the degree of privacy preference that can be controlled by each party. A lot of research has tried to protect differential privacy at the data level when a model is learned in a centralized manner \citep{shokri2015privacy, abadi2016deep}. To protect the data from inversion attack, such as inferring data from model weights,  a differential privacy-preserving randomized mechanism can be incorporated into the learning process.  Given a deterministic real-valued function $h: \mathcal{D} \rightarrow \mathbb{R}^m$, $h$'s $L1$ sensitivity $s_h$  is defined as  the maximum of the absolute distance $\parallel h(\mathcal{D})-h(\mathcal{D'})\parallel_1$, where $\parallel \mathcal{D}-\mathcal{D}'\parallel_1 = 1$, meaning that there is only one data point difference between $\mathcal{D}$ and $\mathcal{D}'$\citep{dwork2014algorithmic} (\textit{Definition 3.1}). In our case $h$ computes the $m$ weight parameters in the deep learning model. Introducing “noise” in the training process (inputs, parameters, or outputs) can limit the granularity of information shared and ensure
 $\epsilon$-differential privacy \citep{dwork2006calibrating} (\textit{Definition 1}) for all $S \subseteq Range(h)$, and then \citep{dwork2006our}:
 \begin{equation}
 Pr[h(\mathcal{D}) \in S ]\leq e^{\epsilon}Pr[h(\mathcal{D}') \in S] ,
 \end{equation} or
 \begin{equation}
 Pr[h(\mathcal{D}) \in S ]\leq e^{\epsilon}Pr[h(\mathcal{D}') \in S] +\delta,
 \end{equation}
 where the additional additive term $\delta$ is the probability of  $\epsilon$-differential privacy being broken.
Here, we introduce two approaches: 1) Gaussian mechanism, and 2) Laplace mechanism, which can enjoy good privacy guarantees \citep{chaudhuri2019capacity} by adding noise to the shared weights. 
\subsubsection*{\textbf{Gaussian Mechanism}} The Gaussian mechanism adds $N(0, s_h^2\sigma^2
)$ noise with mean 0 and standard deviation $s_h\sigma$ to a function $h(\mathcal{D})$ with global sensitivity $s_h$. $h(\mathcal{D})$ will satisfy $(\epsilon,\delta)$-differential privacy if $\delta\geq\frac{4}{5}\text{exp}(-(\sigma\epsilon)^2/2)$ and $\epsilon<1$ \citep{dwork2014algorithmic} (Theorem 3.22). Hereby, we linked the Gaussian noise parameter $\sigma$ to the privacy parameters $\epsilon$ and $\delta$. %The scale of the noise $\alpha$ will be calibrated to $s_h\ln(1/\delta)/\epsilon$. 

\subsubsection*{\textbf{Laplace Mechanism}} The Laplace Distribution centered at 0 with scale $b$ is the distribution with probability density function:
\begin{equation}
   Lap(b):= Lap(x|b) = \frac{1}{2}\text{exp}(-\frac{|x|}{b}),
\end{equation}
and the variance of the Laplace distribution is $\sigma^2 = 2b^2$. The Laplace mechanism adds $Lap(s_h/\epsilon)$ noise to a function $h(\mathcal{D})$ with global sensitivity $s_h$ and preserves $(\epsilon,0)$-difference privacy. Hereby, we linked the Laplace noise parameter $b$ to the privacy parameters $\epsilon$. %The scale of the noise $\alpha$ will be calibrated to $s_h/\epsilon.$

In our case, mapping function $h$ is a deep learning model and it is not tractable to compute the sensitivity $s_h$. For simplicity of discussion, sensitivity $s_h$ is assumed to be 1. From the mechanisms described above, we can control noise parameters to meet certain privacy requirement, as the noise parameters are linked to privacy parameters as shown above.

\subsection{Boosting multi-site learning with domain adaptation}
\label{sec:adaptation}
% \begin{figure*}[t]
%     \centering
%     \includegraphics[width=0.8\textwidth]{figures/adaptation.png}
%     \caption{Domain adaptation strategies for our proposed federated learning setup.}
%     \label{fig:adaptation}
% \end{figure*}

\begin{figure*}[t]
    \centering
    \begin{subfigure}[t]{0.45\textwidth}
        \centering
        \includegraphics[width=0.9\textwidth]{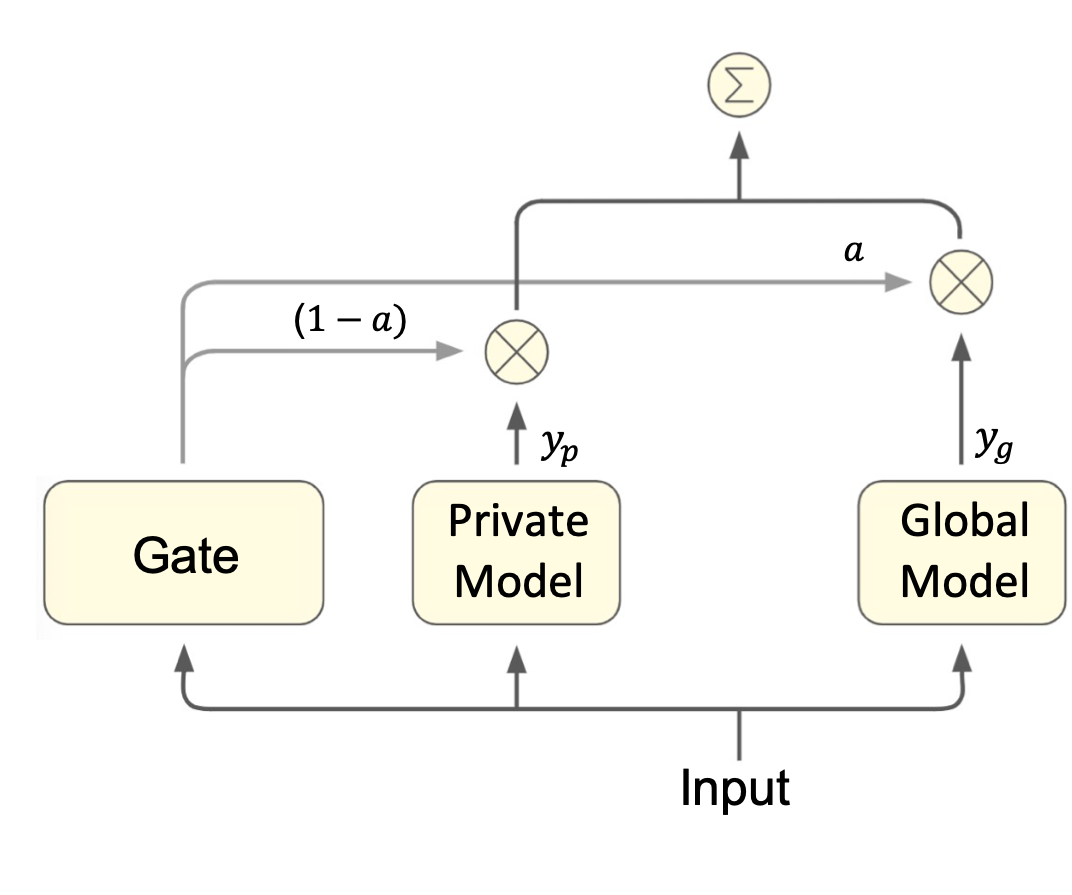}
        \caption{MoE strategy.}
         \label{fig:adaptation-moe}
    \end{subfigure}%
    ~
    \begin{subfigure}[t]{0.45\textwidth}
        \centering
        \includegraphics[width=0.9\textwidth]{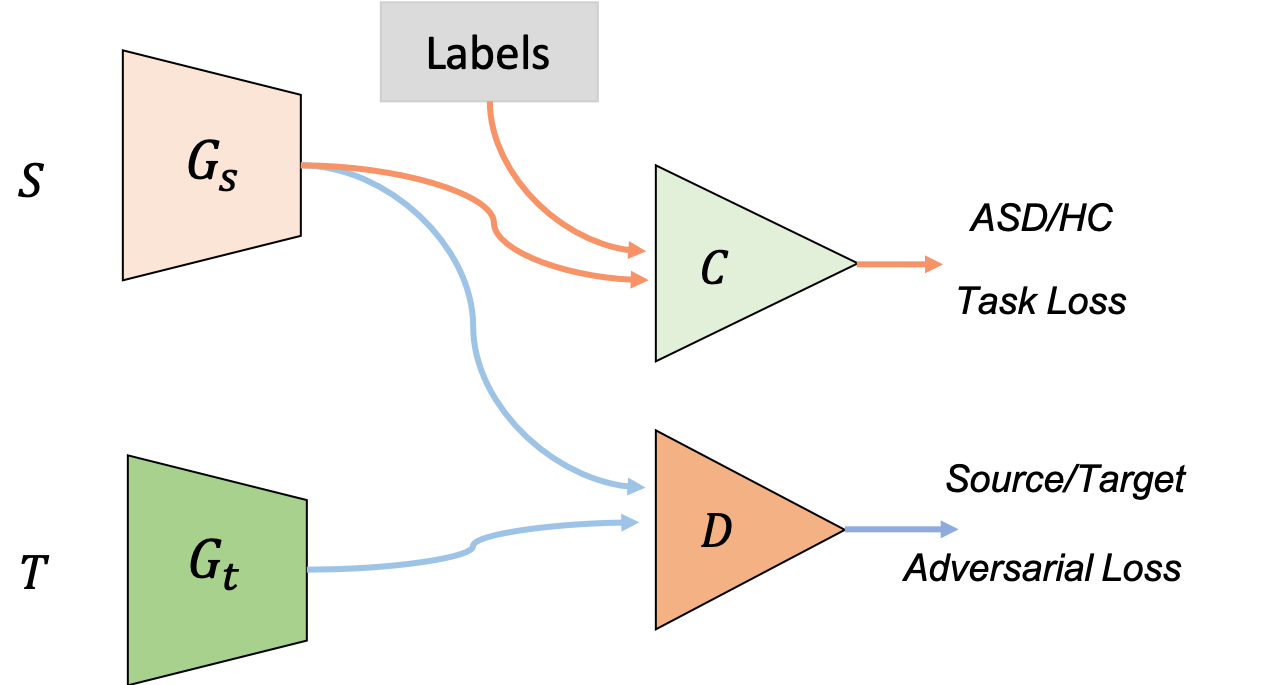}
        \caption{Adversarial alignment strategy}
        \label{fig:adaptation-align}
    \end{subfigure}
    \caption{Domain adaptation strategies for our proposed federated learning setup.}
\end{figure*}

Although federated learning is promising for better privacy and efficiency, there is the additional issue that the data at each site likely have different distributions, leading to domain shift between
the sites \citep{quionero2009dataset}. The main hypothesis here is that domain adaptation techniques can improve overall accuracy of different sites in a federated learning setting, and that holds even when noise is added for privacy-preservation, especially for the sites whose data distributions are quite different from the other sites. In this subsection, we investigate two domain adaptation methods: 1) Mixture of Experts (MoE), adaptation near the output layer, and 2) Adversarial domain alignment, adaptation on the data knowledge representation level. 

\subsubsection{Mixture of Experts (MoE) domain adaptation}
Mixture of Experts (MoE) \citep{masoudnia2014mixture,shazeer2017outrageously,wang2018deep} is an approach to conditionally combine experts to process each input. In deep learning, experts mean deep learning models. An MoE layer for feed-forward neural networks is a trainable gating network that dynamically assigns gated weights to combine multiple networks. Then, all parts of the big model that contains all expert models and the MoE layer are trained jointly by back-propagation.

Mixing the outputs of a collaboratively-learned general model and a domain expert was proposed for domain adaptation \citep{peterson2019private}. Improving from the previous work \citep{peterson2019private}, we integrate randomized mechanism into MoE. Each participating party has an independent set of labeled training examples that they wish to keep private, drawn from a party-specific domain distribution. These users collaborate to build a general model for the task but maintain private, domain-adapted expert models. The final predictor is a weighted average of the outputs from the general and private models. These weights are learned using a MoE architecture \citep{masoudnia2014mixture}, so the entire model can be trained with gradient descent. More specifically, given an input data $x \in X_i$, the output of the global model is $y_G = f_{\bar{w}}(x)$,which is learned using the strategy in Algorithm \ref{ag1}. In the binary classification setting, the output is the model's predicted probability for the positive class. As shown in Figure \ref{fig:adaptation-moe}, we train another local model $f_{\phi_i}$ in the meantime, which is defined as a private model. The private model can have different architecture from $f_{\bar{w}}$ and it does not communicate with the global model.  The output of the private model is $y_P = f_{\phi_i}(x)$. $f_{\phi_i}$ is trained using the regular deep learning setting, without including privacy-related noise.

The final output that entity $i$ uses to label data is
\begin{equation} \label{ensemble}
    \hat{y}_i = a_i(x)y_G + (1- a_i(x))y_P.
\end{equation}
The weight $a_i(x)$ is called a gating function in the MoE, and we use a non-linear layer $a_i(x) = \sigma(\psi_i^T\cdot x +b_i)$ to compute $a_i(x)$, where $\sigma$ is the sigmoid function, and $\psi_i$ and $b_i$ are learned weights by end-to-end training together with the federated learning architecture.

\subsubsection{Adversarial domain alignment}
\begin{algorithm}[htpb]
    		\caption{Federated Adversarial Domain Alignment}\label{ag2}
    		\hspace*{\algorithmicindent} \textbf{Input:} {1. $\mathbf{X}=\{X_1,\dots,X_N\}$, fMRI data from $N$ institutions/sites;  2. $\mathbf{G}_{\theta_{G}}=\{G_{\theta_{G_1}},\dots,G_{\theta_{G_N}}\}$, local feature generators within $N$ sites, where $\theta_{G_n}$ is the generator's parameters of site $n$; 3. $\mathbf{C}_{\theta_{C}}=\{C_{\theta_{C_1}},\dots,C_{\theta_{C_N}}\}$, local classifiers within $N$ sites, where $\theta_{C_n}$ is the classifier's parameters of site $n$;  4. $\mathbf{D}_{\theta_{D}}=\{D_{\theta_{D_1}},\dots,D_{\theta_{D_N}}\}$, discriminators from embedded features, where $\theta_{D_{n}}$ is the discriminator parameters that identify the data from site $n$; 5. $\mathbf{Y} =\{Y_1,\dots,Y_N\}$, fMRI labels (HC or ASD); 6. $M(\cdot)$, noise generator; 7. $K$, number of optimization iterations; 8. $\tau$, global model updating pace; 9. $\{\mathbf{G}_{\bar{\theta}_{G}},\mathbf{C}_{\bar{\theta}_{c}}\}$, global model.
    			%\hspace*{\algorithmicindent} \textbf{Output}
    			\begin{algorithmic}[1]
    				\State {Initialize parameters $\{\theta_G, \theta_C, \theta_D \}$}
    				\For{$k = 1$ to $K$}
    				\State{$t \gets 0$}
    				\Comment{initialize pace counter}
    				\For{$i = 1$ to $N$}
    				\State{Sample mini-batch from source site $\{(X_i^S,Y_i^S)\}_{i=1}^N$ and target site $\{(X_j^T)\}_{j=1}^N$}
    				\State{Compute gradient with cross-entropy classification loss $\mathcal{L}_{ce}$ (Eq. \ref{eq:celoss}) to update $\theta_{G_i}^{(k)}$ and $\theta_{C_i}^{(k)}$}
    				\State{\textbf{Domain Alignment:}}
    				\State{Update $\theta_{D_{i}}^{(k)}, \{\theta_{G_i}^{(k)},\theta_{G_j}^{(k)}\}$ with Eq. \ref{eq:align1} and Eq. \ref{eq:align2} respectively to align the domain distribution}
    				\EndFor
    				\State{$t \gets t+1$}
    				\Comment{models communicate}
    				\If{$t \% \tau = 0$}
    				\State{$\bar{\theta}_G^{(k)} \gets \frac{1}{N}\sum_n (\theta_{G_i}^{(k)}+ M(\theta_{G_i}^{(k)})) $}
    				\State{$\bar{\theta}_C^{(k)} \gets \frac{1}{N}\sum_n (\theta_{C_i}^{(k)}+ M(\theta_{C_i}^{(k)})) $}
    				\Comment{update global model per $\tau$ steps}
    				\For{$n = 1$ to $N$}
    				\State{$\theta_{G_i}^{(k)} \gets \bar{\theta}_G^{(k)}$}
    				\State{$\theta_{C_i}^{(k)} \gets \bar{\theta}_C^{(k)}$}
    				\Comment{deploy weights to local model }
    				\EndFor
    				\EndIf
    				\EndFor	
    				%\State{$\mathcal{M}_{ij} \gets \mathcal{M}_{ij}\mathcal{G}_{ij}$}
    				% \State{$\mathcal{M}_{ij} \gets \frac{\textit{exp}(\mathcal{M}_{ij}A_{ij})}{\sum_{j}\textit{exp}(\mathcal{M}_{ij}A_{ij})}$}
    				% \Comment{calculate probability mask}				
    			\end{algorithmic}
    			\hspace*{\algorithmicindent} \textbf{Return: global model} $\{\mathbf{G}_{\bar{\theta}_{G}},\mathbf{C}_{\bar{\theta}_{c}}\}$
    		}
\end{algorithm}
In the federated setting, the data are locally stored in a privacy-preserving manner. For the domain adaptation problem, we have multiple source domains and want to generalize the domains into a common space of target data.  Due to the data sharing limitation of federated learning, we cannot train a single model that has access to the source domain and target domain simultaneously. To address this issue, we employed federated adversarial alignment \citep{peng2019federated} that introduces two modules (a domain-specific local
feature extractor, and a global discriminator)  in the classification networks and divides optimization into two independent steps. Using this method (Figure \ref{fig:adaptation-align}), for source site  $\mathcal{D}_s$, we train a local feature extractor, $G_s$.  For the target site  $\mathcal{D}_t$, we train a local feature generator $G_t$. For each $(\mathcal{D}_s, \mathcal{D}_t)$ source-target domain pair, we train an adversarial domain discriminator $D$ to align the distributions. First, domain discriminator $D$ is trained to identify which domain the features come from, then the feature generators $(G_s
, G_t)$ are trained to confuse the discriminator $D$. In this setting of privacy preserving, the discriminator $D$ only gets access to the output features with noise coverage of $G_s$ and $G_t$, without leaking the original data. Specifically, the inputs of source discriminator $D$ are $M\circ G_t(x^t)$ and $M\circ G_s(x^s)$, where $M(\cdot)$ is a noise generator. Therefore, the data-leakage of target site is prevented in training the discriminator $D$ on the source side. Given the source domain data $\mathbf{X}^S$ and target data $\mathbf{X}^T$, the objective for discriminating the source domain from the others $D_s$ is defined as:

\begin{equation} \label{eq:align1}
\begin{split}
     \mathop{\mathcal{L}_{advD}}(\mathbf{X}^S,\mathbf{X}^T,G_s,G_t) = &- \mathbb{E}_{x^s \sim \mathbf{X}^S}[\log D_s(G_s(x^s))] \\
     &-\mathbb{E}_{x^t \sim \mathbf{X}^T}[\log (1-D_s(M\circ G_t(x^t)))].
\end{split}
\end{equation}
In the second step, $L_{advD}$ remains unchanged, but  $L_{advG}$ is updated with the following objective:
\begin{equation} \label{eq:align2}
\begin{split} 
     \mathop{\mathcal{L}_{advG}}(\mathbf{X}^S,\mathbf{X}^T,G_s,G_t) = &- \mathbb{E}_{x^s \sim \mathbf{X}^S}[\log D_s(G_s(x^s))] \\
     &-\mathbb{E}_{x^t \sim \mathbf{X}^T}[\log (D_s(M\circ G_t(x^t))].
\end{split}
\end{equation}
By end-to-end training of the federated learning model with the alignment module, we can minimize the discrepancy
between the source and target domains. The implementation details are described in Algorithm \ref{ag2}.

\subsection{Evaluate model by interpreting biomarkers}
\label{sec:evalinterp}
 The primary goal of psychiatric neuroimaging research is to identify objective and repeatable biomarkers that may inform the disease \citep{heinsfeld2018identification}. Finding the biomarkers associated with ASD is extremely helpful in understanding the underlying roots of the disorder and can lead to earlier diagnosis and more targeted treatment. Alteration in brain functional connectivity is expected to provide potential biomarkers for classifying or predicting brain disorders \citep{du2018classification}. Deep learning methods are promising tools for investigating the reliability of patterns of brain function across large and heterogeneous data sets \citep{varoquaux2014machine}.
 
 We held the hypothesis that reliable biomarkers could be detected from a reliable model. The guided gradient-based explanation method \citep{simonyan2013deep, springenberg2014striving} is perhaps the most straightforward and easiest approach for data feature importance interpretation. The advantage of gradient-based explanation method is easy to compute. By calculating the difference of the output w.r.t the model input then applying norm, a score can be obtained. The gradient-based score can be used to indicate the relative importance of the input feature since it represents the change in input space, which corresponds to the positive maximizing rate of change in the model output. 
\begin{equation} \label{eq:grad}
	g_{k}^c = \textit{ReLU} \left(\frac{\partial \hat{y}^c}{\partial x_{k}} \right)
\end{equation}
where $c \in \{0,\ldots, \mathcal{C}-1\}$ is the correct class of input, $\mathcal{C}$ is the total number of classes, and $y^c$ is the score for class $c$ before softmax layer, $x_k$ is the $k$th feature of the input. $g_{k}^c$ can indicate the importance of feature $k$ for classifying an input as class $c$. We use this method to interpret the important features (ROIs) as biomarkers. 

{Given the important biomarkers, first, we propose to examine their consistency, i.e., whether the biomarkers are robust across different datasets. Second, we should examine whether the biomarkers are meaningful. For the relatively important features selected, such as the features with the top $K$ important scores, we can "decode" them to associated functional keywords based on prior knowledge and compute the correlation score $v_{keyword}^c$ for the keyword with the biomarkers in class $c$. The informative biomarkers of the inputs in the different classes $c$ should have different functional representations, which means we expect large $|\Delta| = |v_{keyword}^c - v_{keyword}^{c'}|$ for the informative biomarkers, where $c' \in \mathcal{C}\setminus c$ . The larger the difference, the more representative and informative the biomarkers.} 
%We normalize $ g^c = [g_k^c]_{k=1}^{6105}$ by dividing  $ max(g^{c})$ to be bound it to $[0,1]$.
\section{Experiments and Results}\label{sec:experiment}
\subsection{Data}
\subsubsection{Participants}
The study was carried out using resting-state fMRI (rs-fMRI) data from the Autism Brain Imaging Data Exchange dataset (ABIDE I preprocessed, \citep{di2014autism}). ABIDE is a consortium that provides preciously collected rs-fMRI ASD and matched controls data for the purpose of data sharing in the scientific community. However, in reality, collecting data in a consortium like ABIDE is not easy as strict agreement need to be reached by different parties. Therefore, although the data were shared in ABIDE, we studied the multi-site data from the federated learning perspective. To ensure the deep learning model could perform on a single site, we downloaded Regions of Interests (ROIs) fMRI series of the top four largest sites (UM1, NYU, USM, UCLA1) from the preprocessed ABIDE dataset with Configurable Pipeline for the Analysis of Connectomes (CPAC), band-pass filtering (0.01 - 0.1 Hz), no global signal regression, parcellated by Harvard-Oxford (HO) atlas. Skipping subjects lacking filename, we downloaded 106, 175, 72, 71 subjects from UM1, NYU, USM, UCLA1 separately. HO parcellated each brain into 111 ROIs. Since some subjects did not contain complete ROIs, we removed the incomplete data, resulting in 88, 167, 52, 63 subjects for UM1, NYU, USM, UCLA1 separately. Due to a lack of sufficient data, we used sliding windows (with window size 32 and stride 1) to truncate raw time sequences of fMRI. After removing incomplete subjects, the compositions of four sites were
shown in Table \ref{tab:summary}. We denoted UM for UM1 and UCLA for UCLA1. We summarized the phenotype information of the subjects under our study in Table \ref{tab:phenotype}.
\begin{table}[htpb]
\centering
\begin{tabular}{@{}l|cccc@{}}
\toprule
% \multicolumn{5}{c}{\textbf{Data Summary}} \\
% \hline
 & \textbf{NYU} & \textbf{UM} & \textbf{USM} & \textbf{UCLA} \\
 \hline
Total Subject & 167 & 88 & 52 & 63 \\
ASD Subject & 73 & 43 & 33 & 37 \\
HC Subject & 94 & 45 & 19 & 26 \\
ASD Percentage & 44\% & 49\% & 63\% & 59\% \\
fMRI Frames & 176 & 296 & 236 & 116 \\
Overlapping Trunc & 145 & 265 & 205 & 85\\
\bottomrule
\end{tabular}%

\caption{Data summary of the dataset used in our study}
\label{tab:summary}
\end{table}

\begin{table}[]
\centering
\resizebox{0.45\textwidth}{!}{%
\begin{tabular}{@{}c|lcccl@{}} \toprule
\multicolumn{1}{c}{} & \textbf{SITE} & \multicolumn{1}{c}{\textbf{AGE}} & \multicolumn{1}{c}{\textbf{ADOS}} & \multicolumn{1}{c}{\textbf{IQ}} & \multicolumn{1}{c}{\textbf{SEX}} \\ \midrule
\multirow{4}{*}{ASD} & UM & 12.4(2.2) & - & 102.8(18.8) & M 36 F 7 \\
 & USM & 22.9(7.3) & 12.6(3.0) & 99.8(16.4) & M 33 F 0 \\
 & NYU & 14.7(7.1) & 11.5(4.1) & 107.4(16.5) & M 65 F 8 \\
 & UCLA & 13.0(2.7) & 10.4(3.6) & 103.5(13.5) & M 31 F 6 \\ \midrule
\multirow{4}{*}{HC} & UM & 14.1(3.4) & - & 106.7(9.6) & M 32 F 13 \\
 & USM & 20.8(8.2) & -  & 117.1(14.4) & M 19 F 0 \\
 & NYU & 15.2(5.9) & - & 112.6(13.5) & M 69 F 25 \\
 & UCLA & 13.4(2.3) & -  & 104.9(10.4) & M 22 F 4 \\ \bottomrule
\end{tabular}%
}
Values reported with mean (std) format. M: Male, F: Female, ADOS score: - means information not available
\caption{Data phenotype summary.}
\label{tab:phenotype}
\end{table}

\subsubsection{Data preprocessing}
The task we performed on the ABIDE datasets was to identify autism spectrum disorders (ASD) or healthy control (HC). We used the mean time sequences of ROIs to compute the correlation matrix as functional connectivity. The functional connectivity provided an index of the level of co-activation of brain regions based on the time series of rs-fMRI brain imaging data. Each element of the correlation matrix was calculated using Pearson correlation coefficient, which ranged from -1 to 1: values close to 1 indicated that the time series were highly correlated and values close to -1 indicate that the time series are anti-correlated. Then, we applied the Fisher transformation on the correlation matrices to emphasize the strong correlations. As the correlation matrices were symmetric, we only kept the upper-triangle of the matrices and flattened the triangle values to vectors, with the purpose of using them for the inputs of multilayer perceptron (MLP) classifiers. The number of resultant features was defined by $R(R-1)/2$, where $R$ was the number of ROIs. Under the HO atlas (111 ROIs), the procedure resulted in 6105 features. 

\subsection{Federated training setup and hyper-parameters discussion}
\label{sec:fedtrain}
A multi-layer perceptron (MLP) 6105-16-2 (corresponding to 6105 nodes for the input (first) layer, 16 nodes for the hidden layer, and 2 nodes for the output layer) was used for classification. The outputs of the MLPs were the probability of the given input being classified as each class. We used cross-entropy as the objective function. We performed 5-fold cross-validation (subject-wise splitting), and each entry of the input vectors was normalized by training set mean and standard deviation (std) within each site. As we performed overlapping truncation for data augmentation in data processing, we used the majority voting method to evaluate the final classification performance. For example, we augmented $m$ input instances for a single subject, and if more than $m/2$ instances were classified as ASD, then we assigned 'ASD' label to the subject. Adam optimization was applied with initial learning rate 1e-5 and reduced by 1/2 for every 20 epochs and stopped at the 50th epoch. In each epoch, we performed local updates multiple times instead of once based on communication pace $\tau$. We set the total steps of each epoch as 60, and the batch size of each site was the number of training data over 60. 

\begin{figure}[htpb]
    \centering
    \includegraphics[width=0.45\textwidth]{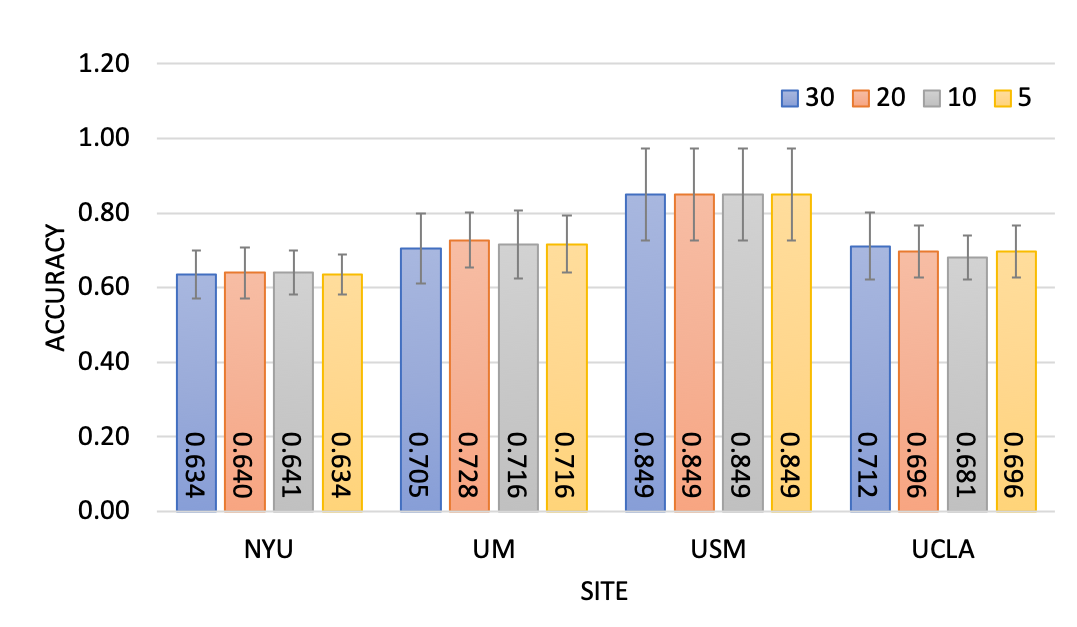}
    \caption{Investigate communication pace $\tau$ vs accuracy}
    \label{fig:pace}
\end{figure}

First, we investigated the effects of changing communication pace on classification accuracy, as communication between models would be costly. To select the best communication pace $\tau$, we did not apply any noise on the shared weights in the experiment. As the results in Figure \ref{fig:pace} show, there was no significant difference between the accuracies when $\tau$ varied from 5 to 30.

\begin{figure}[htpb]
    \centering
    \includegraphics[width=0.45\textwidth]{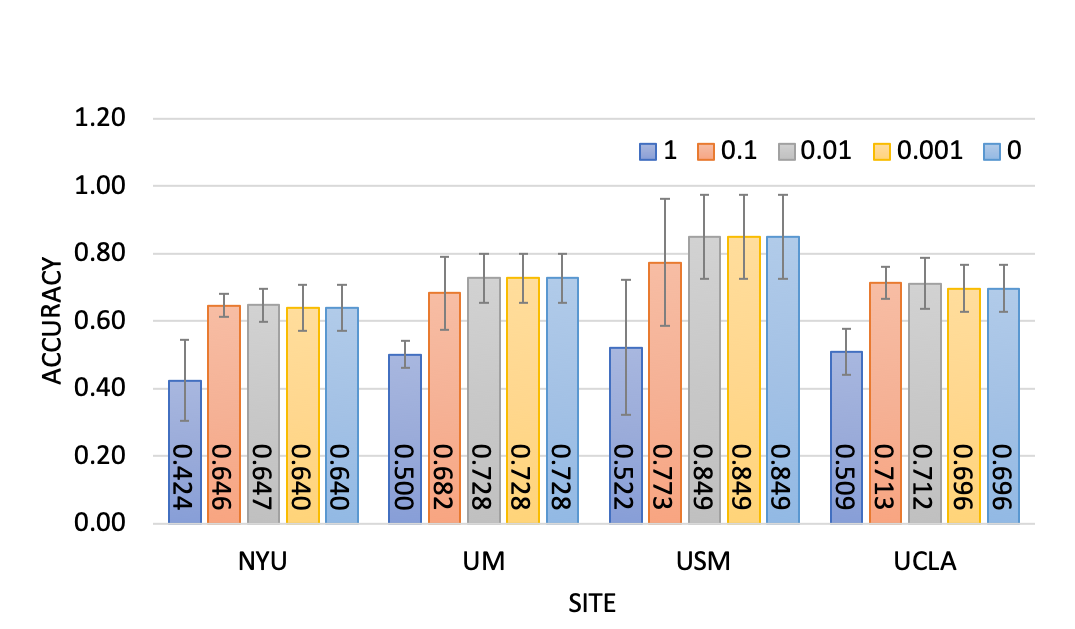}
    \caption{Investigate Gaussian mechanism vs accuracy}
    \label{fig:gau}
\end{figure}

\begin{figure}[htpb]
    \centering
    \includegraphics[width=0.45\textwidth]{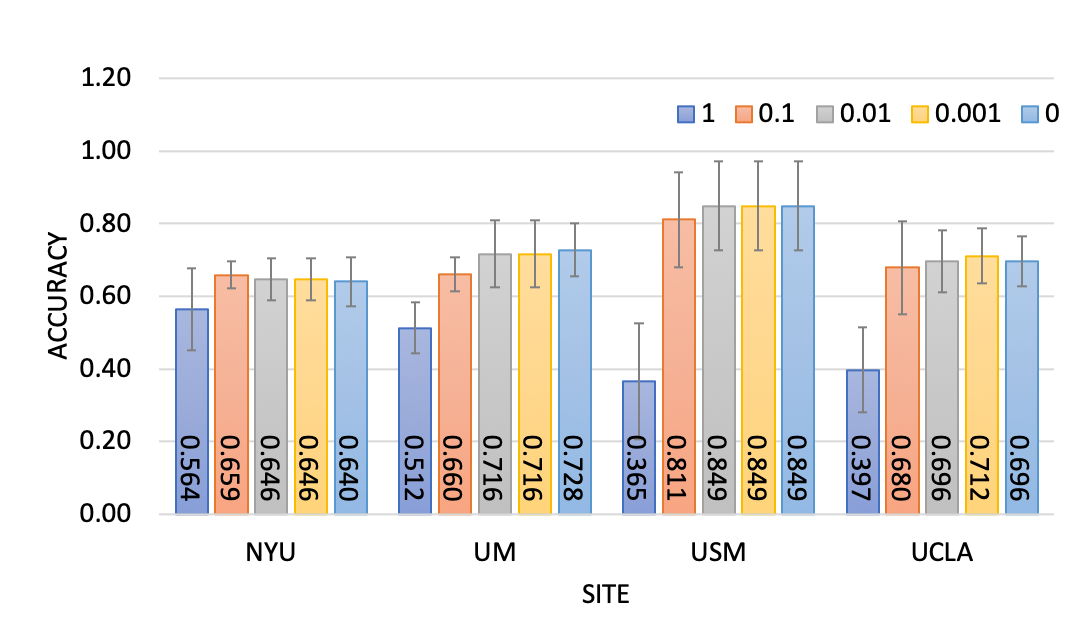}
    \caption{Investigate Laplace mechanism vs accuracy}
    \label{fig:lap}
\end{figure}

Then, we investigated adding the randomization mechanism on shared weights to protect the data from inversion attack, such as inferring data from model weights, given local model weights. Here we tested the Gaussian and Laplace mechanism, which corresponded to L2 and L1 sensitivity. Institutions may want to specify the level of privacy they want to preserve, which would be reflected in the noise levels. For the Gaussian mechanism experiment, we generated Gaussian noise $\varepsilon_n\sim N(0,\alpha\sigma)$  adding to local model weights, where $\sigma$ was the standard deviation of the local model weights and $\alpha$ was the noise level. We varied $\alpha$ from 0.001 to 1. For the Laplace mechanism experiment, we generated Laplace noise $\varepsilon_n\sim Lap(\alpha\sigma/\sqrt{2})$ adding to local model weights, where $\alpha$ was the scale parameter, and $\sigma$ was the standard deviation of the local model weights. We varied $\alpha$ from 0.001 to 1. As the results in Figure \ref{fig:gau} and Figure \ref{fig:lap} show, there was a trade-off between model performance and noise level (privacy-preserving level). When the noise level was too high ($\alpha = 1$ in our setup), corresponding to high privacy-preserving levels, the models failed in the classification task.
%Hence, it provided strong protection against indirect data leakage.  
\subsection{Comparisons with different strategies}
\label{sec:startegyres}
\begin{figure*}[t]
    \centering
    \includegraphics[width=0.9\textwidth]{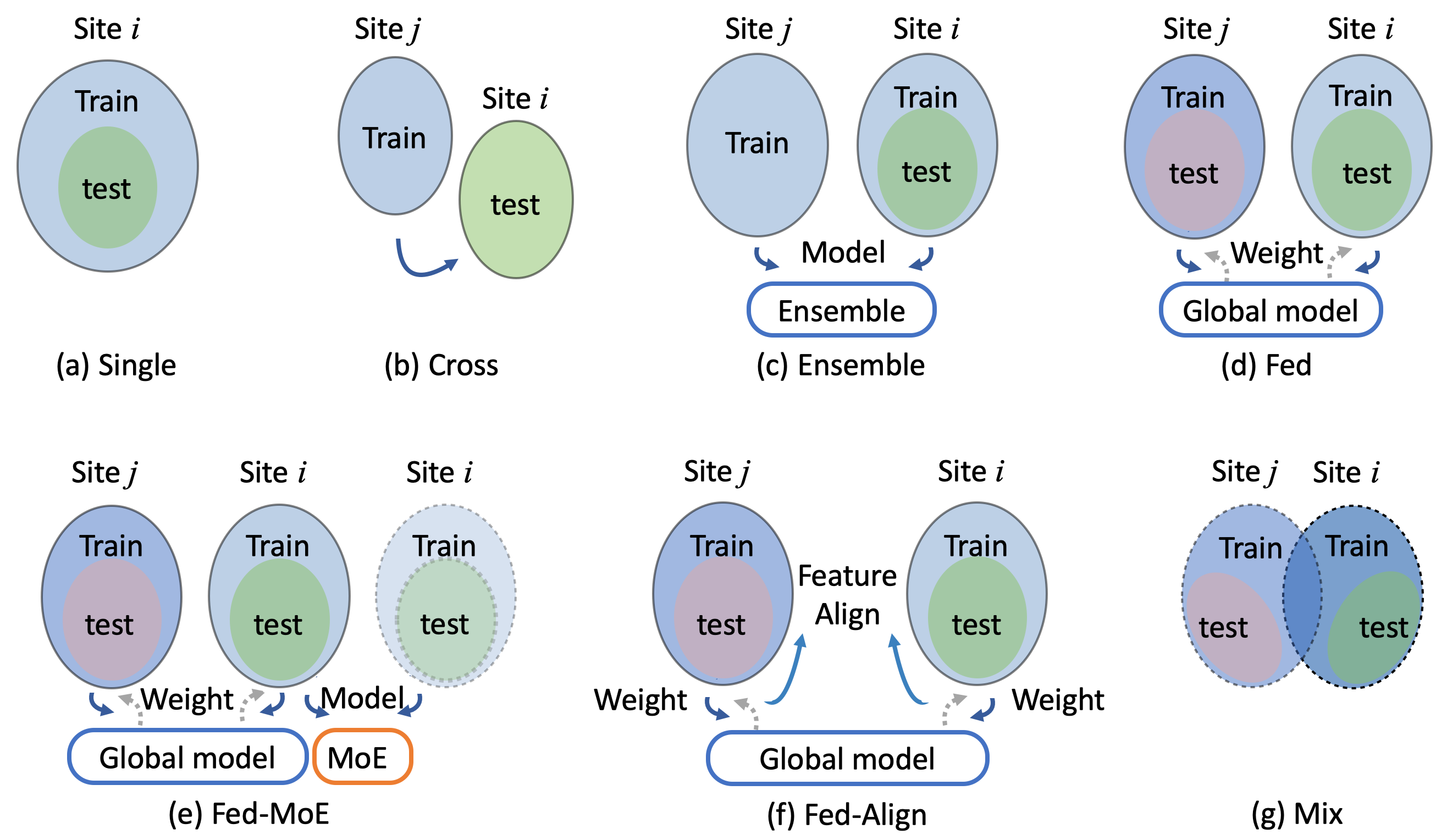}
    \caption{Different classification strategies}
    \label{fig:strategy}
\end{figure*}
To demonstrate the proposed federated learning framework in Algorithm \ref{ag1} (\textit{Fed}) could improve multi-site fMRI classification, we compared the proposed methods ($\tau = 20$ and $\alpha=0.01$) with four alternative, non-federated strategies: 1) training and testing within the single site (\textit{Single}); 2) training using one site and testing on another site (\textit{Cross}); 3) collecting multi-site data together for training (\textit{Mix}) and 4) creating an ensemble model using the models from different sites (\textit{Ensemble}). \textit{Ensemble} method averaged the outputs from a\textit{Single} model that was trained within the site and a \textit{Cross} model that was trained by another site. \textit{Single} and \textit{Cross} preserved data privacy, while could not incorporate the data. \textit{Mix} could take use of all the data from different sites, while could not preserve data privacy. The classification performance of \textit{Mix} was expected to perform better than \textit{Fed} as it used more data information. To fairly compare the results, we tried to choose the best model parameters for different strategies by varying the model as little as possible. Because the sizes of the available training data for \textit{Single}, \textit{Cross} and \textit{Ensemble} strategies were much smaller than those of \textit{Mix} and \textit{Fed} strategies, we changed the MLP architecture to 6105-8-2 to avoid overfitting, while all the other training settings and data splitting settings were the same as described in Section \ref{sec:fedtrain}.

Considering the fact that data distribution was heterogeneous, we also tried to use the domain adaptation methods introduced in Section \ref{sec:adaptation} to boost the classification performance of \textit{Fed}. For the combination of federated training and MoE (\textit{Fed-MoE}) strategy, we trained a private classifier simultaneously with the federated architecture. The same as \textit{Single}, we used MLPs 6105-8-2 as the private models. The gate function was implemented by an MLP with two fully-connected (FC) layers 6105-8-1 and a sigmoid non-linearity layer. For the combination of federated training and adversarial alignment (\textit{Fed-Align}) strategy, we used four discriminators $D$ to discriminate whether the data came from the source domain. We treated the first two layers of the federated MLPs 6105-16 as a feature generator $G$, and each site had a different $G$. Following the randomized mechanism in \textit{Fed}, we sent the generated features blurred by Gaussian noises $\varepsilon_n\sim N(0,0.01\sigma)$ to the inputs of $D$. The input of the classifier $C$ was a 16-dim vector. The global model was the concatenation of $G-C$. Only the $G$ and $C$ weights of local models were shared with the global model. For the whole network training, the setup was the same as training a \textit{Fed} model, except that we started to propagate adversarial loss on $D$ (Eq. \ref{eq:align1}) after training the $G-C$ part for 5 epochs.

How to utilize data for training and testing in different classification strategies was explained in Figure \ref{fig:strategy}. All the implemented model architectures were shown in the Appendix. The comparison results were shown in Table \ref{tab:strategy}. In \textit{Cross}, we denoted the site used for training as 'tr$<$site$>$'. As the testing data were all the other whole sites, there was no standard deviation (std) to report. Also, we ignored the performance of the site used for training. The other results were reported using the 'mean (std)' format. By comparing the mean accuracy only, we highlighted the best accuracy in Table \ref{tab:strategy}. The reason why \textit{Cross} results were better than \textit{Single} was probably because more data were included in training (no data splitting). For example, the total number of training instances at the UCLA site with \textit{Single} strategy was $85\times63\times0.8 \text{ (5-fold)}= 4284$, while using the \textit{Cross} strategy training on the USM site then testing on the UCLA site included $205\times52=10660$ training instances. \textit{Ensemble} results were not good, probably because the ensemble methods could not make use of the decisions made by different models and counter-productively weakened the prediction power. The mean accuracy of \textit{Fed} was higher than the best \textit{Cross} learning case for each single site.
In addition, \textit{Fed} was significantly better than \textit{Single} by two sample t test with $p<0.001$ for each site. We also observed that \textit{Fed-MoE} ($p=0.003$)  and \textit{Fed-Align} ($p<0.001$) significantly improved accuracy on NYU site when comparing with \textit{Fed}. The accuracy on UM site using \textit{Fed-Align} was significantly better than the accuracy using \textit{Fed} ($p=0.018$).The accuracy on UCLA site using \textit{Fed-MoE} showed potential to improve the classification results compared with using \textit{Fed} ($p=0.094$). Using domain adaptation methods did not improve the performance on the USM site, which was probably caused by the data distribution of the USM site. We validated the hypothesis in the following discussion.

\begin{table}[]
\centering
\resizebox{0.45\textwidth}{!}{%
\begin{tabular}{@{}l|cccc@{}} \toprule
       & \textbf{NYU} & \textbf{UM}  & \textbf{USM} & \textbf{UCLA} \\  \hline
trNYU  & -            & 0.716        & 0.673        & 0.682         \\
trUM   & 0.611        & -            & 0.712        & 0.682         \\
trUSM  & 0.641        & 0.625        & -            & 0.730         \\
trUCLA & 0.575        & 0.648        & 0.750        & -             \\
Single & 0.601(0.064) & 0.648(0.065) & 0.695(0.108) & 0.571(0.100)  \\
Ensemble &0.611(0.012)& 0.638(0.054)  & 0.654(0.088)  &  0.634(0.064) \\
Fed    & 0.647(0.049) & 0.728(0.073) & \textbf{0.849(0.124)} & 0.712(0.075)  \\
Fed-MoE    & 0.671(0.082) & 0.728(0.083) & 0.809(0.098) & \textbf{0.744(0.130)}  \\
Fed-Align  & \textbf{0.676(0.071)} & \textbf{0.751(0.053)} & 0.829(0.091) & 0.712(0.089)  \\
Mix    & 0.671(0.035) & 0.740(0.063) & 0.829(0.137) & 0.710(0.128)  \\ \bottomrule
\end{tabular}%
}
\caption{Results of using different training strategies}
\label{tab:strategy}
\end{table}
\subsection{Evaluate model from interpretation perspective}
\label{sec:interpres}

We tried to understand the model mechanism by interpreting how each model made a particular decision and how the adaptation methods affected the decision-making process.
\subsubsection{Aligned feature embedding}
\begin{figure}[htbp]
    \centering
    \begin{subfigure}[t]{0.23\textwidth}
        \centering
        \includegraphics[width=0.9\textwidth]{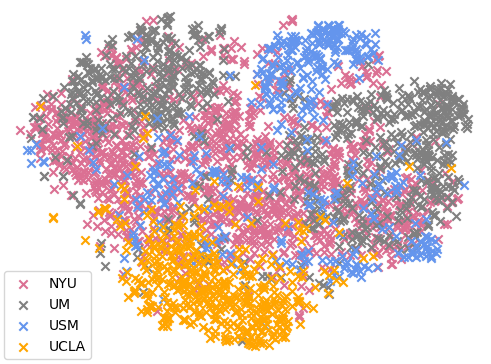}
        \caption{Embedded latent features from 4 sites without alignment.}
        \label{fig:latent1}
    \end{subfigure}%
    ~
    \begin{subfigure}[t]{0.23\textwidth}
        \centering
        \includegraphics[width=0.9\textwidth]{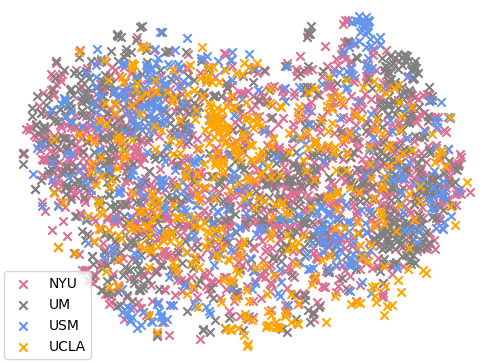}
        \caption{Embedded latent features from 4 sites with alignment.}
        \label{fig:latent2}
    \end{subfigure}
    \caption{t-SNE visualization of latent space.}
    
\end{figure}

We used t-SNE \citep{maaten2008visualizing} to visualize the latent space embedded by the first fully connected layer in Figure \ref{fig:latent1} and Figure \ref{fig:latent2} for our federated learning model without and with adversarial domain alignment. We found the alignment method overall improved domain adaptation.
In Figure \ref{fig:latent1}, we also noticed that the features of the USM site (blue crosses) mixed with other domains. We assumed that could be the reason why the adversarial domain alignment methods did not improve federated learning accuracy for the USM site. 

\subsubsection{MoE gating value}
\begin{figure}[htpb]
    \centering
    \includegraphics[width=0.45\textwidth]{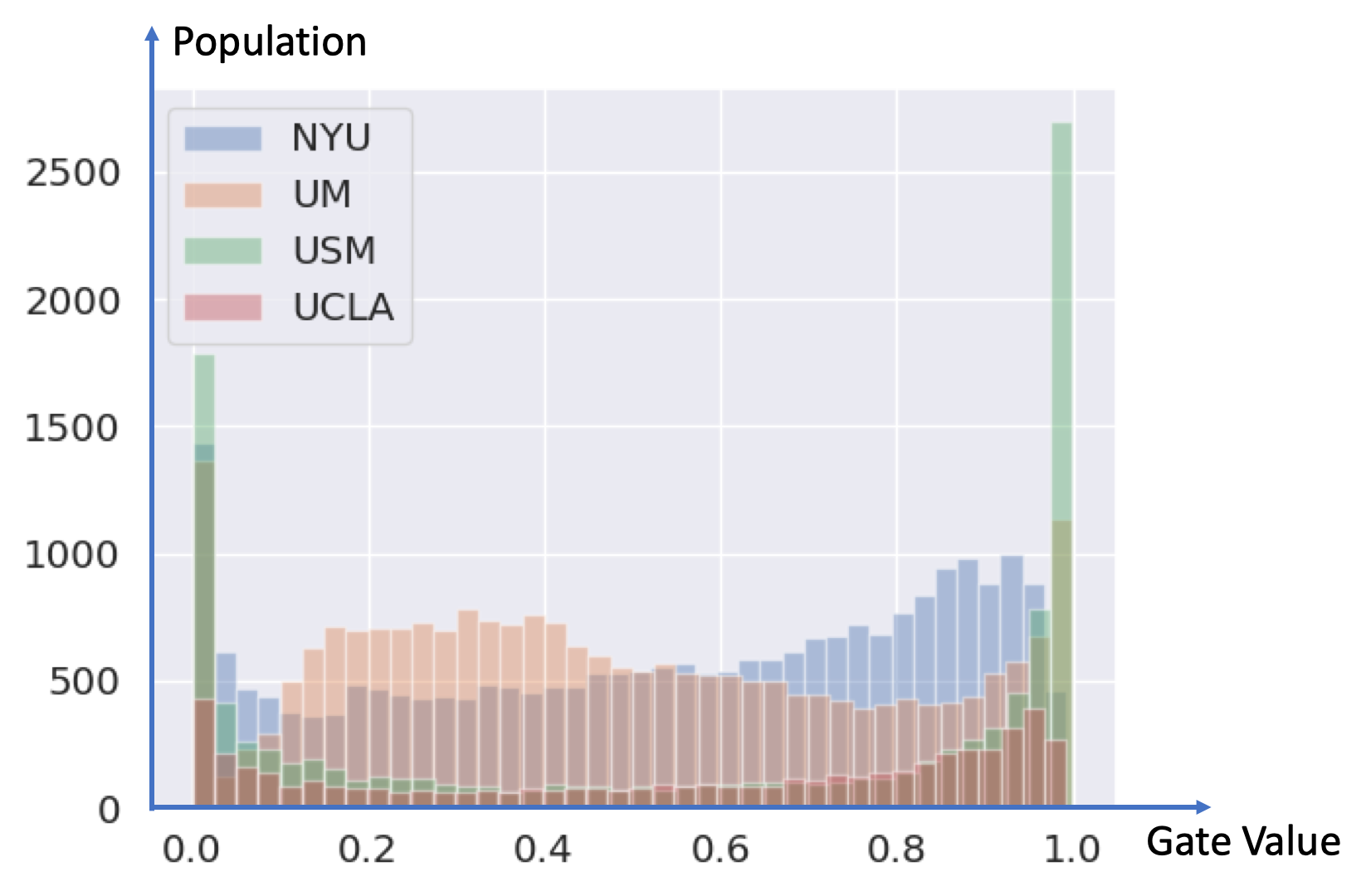}
    \caption{The histogram of MoE gated values assigned to federated global model.}
    \label{fig:gate}
\end{figure}

The core of MoE was to mix the outputs of a collaboratively-learned global model and a private model in each site. Over time, a site's gate function $a(x)$ learned whether to trust the global model or the private model more for a given input. The private model needed to perform well on only the subset of the data points for which the global model failed. While the global model still benefited from the data product (model weights) sharing but received weaker updates on these hard "private" data points. This meant that users with unusual domains had a smaller effect on the global model, which might increase their ability to generalize \citep{ji2019learning}. We show the gating value associated with a federated global model for each testing data point in Figure \ref{fig:gate}. Again, we noticed that the gating values were almost uniformly distributed in the range $[0,1]$, which meant the MoE layer functioned as an inter-medium to coordinate the decisions of the private and global model, except that the gating values of USM site were skewed to 0s and 1s. This showed evidence for why \textit{Fed-MoE} did not perform better than \textit{Fed} on the USM site.

\subsubsection{Neural patterns: connectivity in the autistic brain}
\label{sec:neuropattern}
% \begin{figure}[htpb]
%     \centering
%     \begin{subfigure}[t]{0.25\textwidth}
%         \centering
%         \includegraphics[width=\textwidth]{figures/edge_score.png}
%         \caption{The histogram of Connectivity importance score distribution.}
%         \label{fig:edge}
%     \end{subfigure}%
%     ~
%     \begin{subfigure}[t]{0.25\textwidth}
%         \centering
%         \includegraphics[width=\textwidth]{figures/roi_score.png}
%         \caption{The histogram of ROI importance score distribution}
%         \label{fig:roi}
%     \end{subfigure}
%     \caption{Finding important bio-features associated with ASD}
% \end{figure}

\begin{figure*}[t]
    \centering
    \begin{subfigure}[t]{0.25\textwidth}
        \centering
        \includegraphics[width=0.98\textwidth]{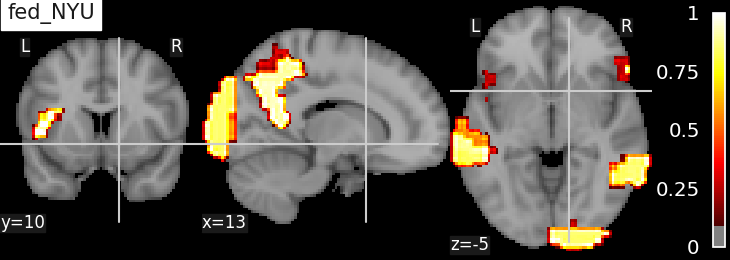}
        % \caption{Detected biomarkers for NYU site using \textbf{\textit{Fed}} strategy - view 1.}
    \end{subfigure}%
    ~
    \begin{subfigure}[t]{0.25\textwidth}
        \centering
        \includegraphics[width=0.98\textwidth]{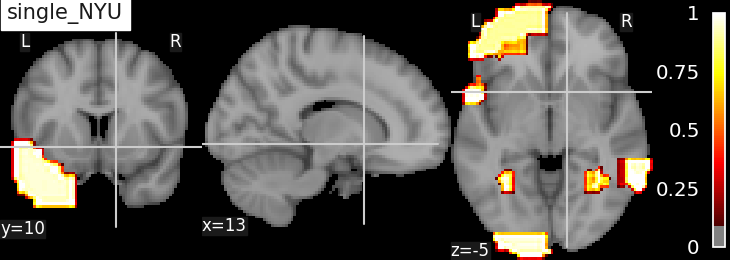}
        % \caption{Detected biomarkers for NYU site using \textbf{\textit{Fed}} strategy - view 2.}
    \end{subfigure}%
    ~
    \begin{subfigure}[t]{0.25\textwidth}
        \centering
        \includegraphics[width=0.98\textwidth]{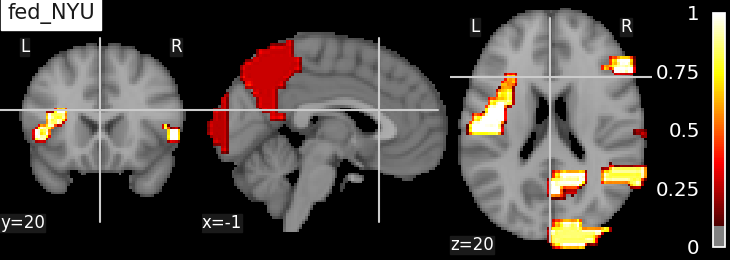}
        % \caption{Detected biomarkers for UM site using \textbf{\textit{Fed}} strategy - view 1.}
    \end{subfigure}%
    ~
    \begin{subfigure}[t]{0.25\textwidth}
        \centering
        \includegraphics[width=0.98\textwidth]{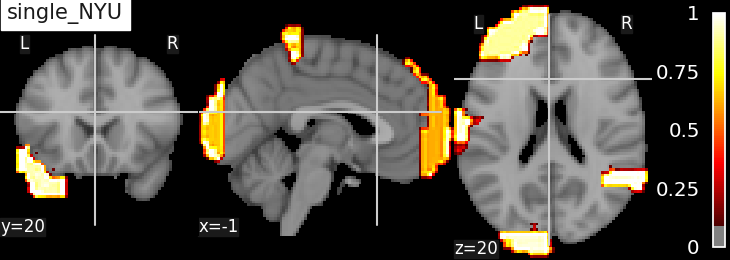}
        % \caption{Detected biomarkers for UM site using \textbf{\textit{Fed}} strategy - view 2.}
    \end{subfigure}%
    ~
    
    \begin{subfigure}[t]{0.25\textwidth}
        \centering
        \includegraphics[width=0.98\textwidth]{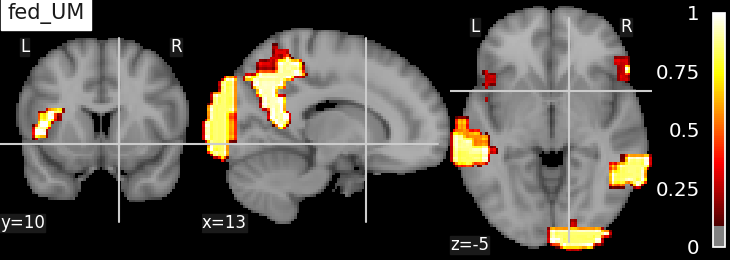}
        % \caption{Detected biomarkers for USM site using \textbf{\textit{Fed}} strategy - view 1.}
    \end{subfigure}%
    ~
    \begin{subfigure}[t]{0.25\textwidth}
        \centering
        \includegraphics[width=0.98\textwidth]{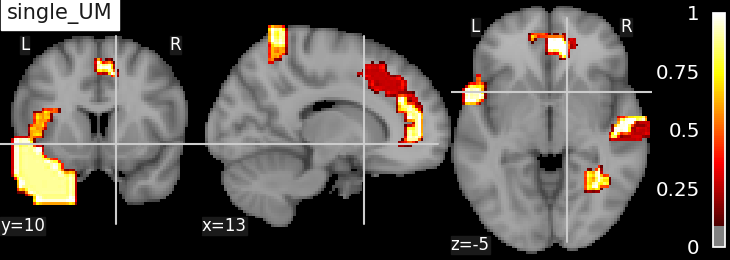}
        % \caption{Detected biomarkers for USM site using \textbf{\textit{Fed}} strategy - view 2.}
    \end{subfigure}%
    ~
    \begin{subfigure}[t]{0.25\textwidth}
        \centering
        \includegraphics[width=0.98\textwidth]{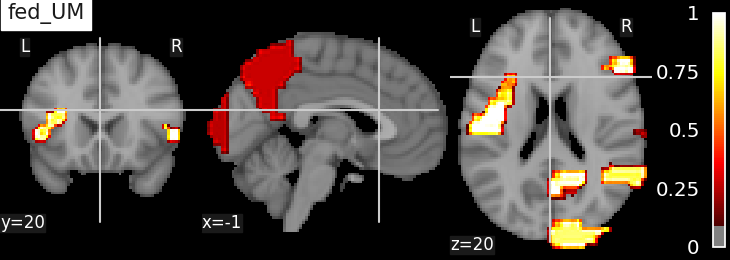}
        % \caption{Detected biomarkers for UCLA site using \textbf{\textit{Fed}} strategy - view 1.}
    \end{subfigure}%
    ~
    \begin{subfigure}[t]{0.25\textwidth}
        \centering
        \includegraphics[width=0.98\textwidth]{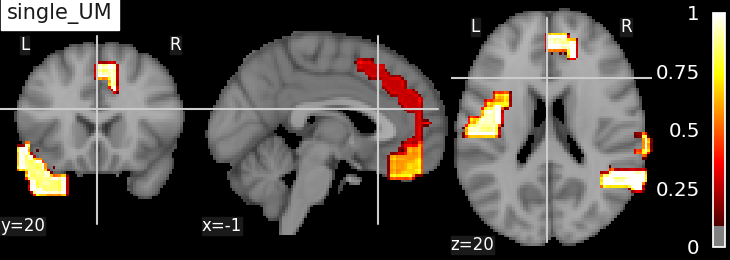}
        % \caption{Detected biomarkers for UCLA site using \textbf{\textit{Fed}} strategy - view 2.}
    \end{subfigure}%
    ~
    % \caption{Interpreting brain biomarkers associated with identifying ASD from federated learning model (\textit{Fed})} 
    % \label{fig:fedbio}
% \end{figure}

% \begin{figure}[t]
    % \centering
    \begin{subfigure}[t]{0.25\textwidth}
        \centering
        \includegraphics[width=0.98\textwidth]{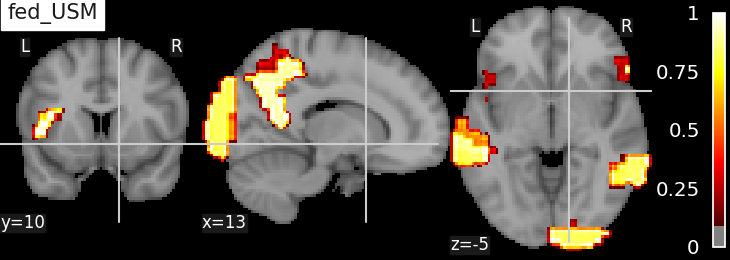}
        % \caption{Detected biomarkers for NYU site using \textbf{\textit{Single}} strategy - view 1.}
    \end{subfigure}%
    ~
    \begin{subfigure}[t]{0.25\textwidth}
        \centering
        \includegraphics[width=0.98\textwidth]{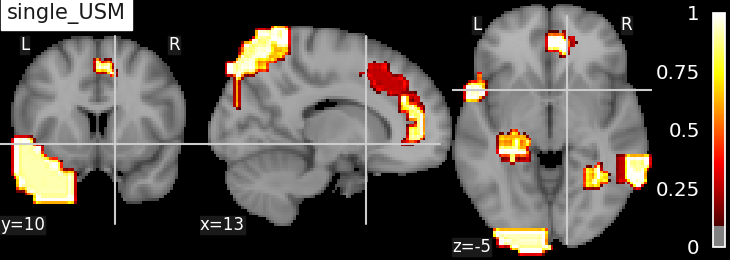}
        % \caption{Detected biomarkers for NYU site using \textbf{\textit{Single}} strategy - view 2.}
    \end{subfigure}%
    ~
    \begin{subfigure}[t]{0.25\textwidth}
        \centering
        \includegraphics[width=0.98\textwidth]{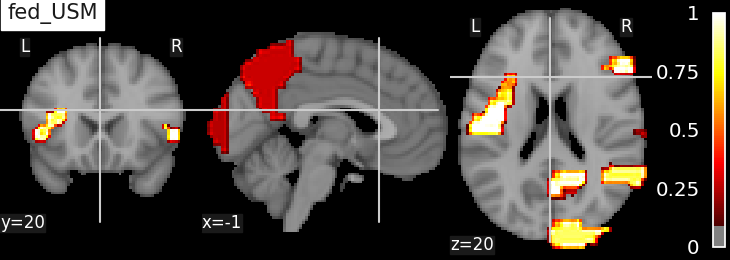}
        % \caption{Detected biomarkers for UM site using \textbf{\textit{Single}} strategy - view 1.}
    \end{subfigure}%
    ~
    \begin{subfigure}[t]{0.25\textwidth}
        \centering
        \includegraphics[width=0.98\textwidth]{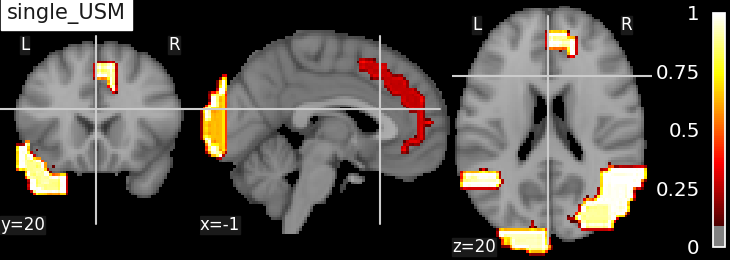}
        % \caption{Detected biomarkers for UM site using \textbf{\textit{Single}} strategy - view 2.}
    \end{subfigure}%
    ~
    
    \begin{subfigure}[t]{0.25\textwidth}
        \centering
        \includegraphics[width=0.98\textwidth]{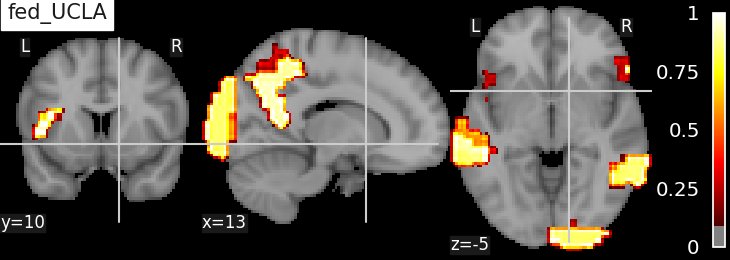}
        \caption{Biomarkers using \textbf{\textit{Fed}} strategy - view 1.}
    \end{subfigure}%
    ~
    \begin{subfigure}[t]{0.25\textwidth}
        \centering
        \includegraphics[width=0.98\textwidth]{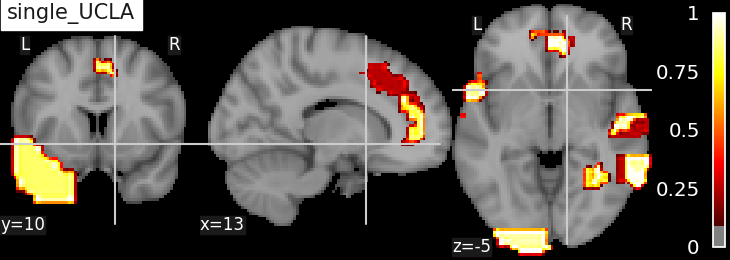}
        \caption{Biomarkers using \textbf{\textit{Single}} strategy - view 1.}
    \end{subfigure}%
    ~
    \begin{subfigure}[t]{0.25\textwidth}
        \centering
        \includegraphics[width=0.98\textwidth]{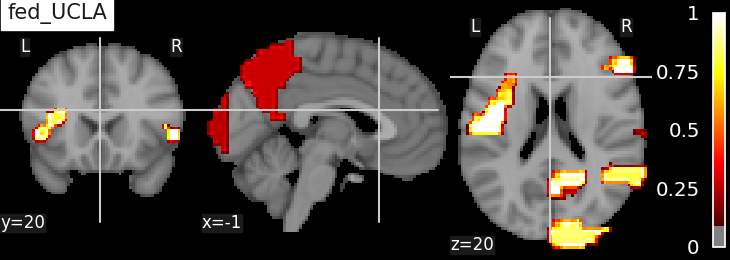}
        \caption{Biomarkers using \textbf{\textit{Fed}} strategy - view 2.}
    \end{subfigure}%
    ~
    \begin{subfigure}[t]{0.25\textwidth}
        \centering
        \includegraphics[width=0.98\textwidth]{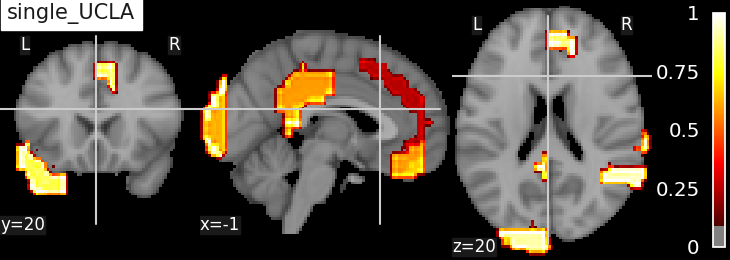}
        \caption{Biomarkers using \textbf{\textit{Single}} strategy - view 2.}
    \end{subfigure}%
    ~
    \caption{Interpreting brain biomarkers associated with identifying \textbf{HC} from federated learning model (\textit{Fed}) and using single site data for training (\textit{Single}). The colors stand for the relative importance scores of the ROIs and the values are denoted on the color bar. The names of the strategies and sites are denoted on the left-upper corners of each subfigure. Each row shows the results of NYU, UM, USM, UCLA site from top to bottom.} 
    \label{fig:hcbio}
\end{figure*}

\begin{figure*}[t]
    \centering
    \begin{subfigure}[t]{0.25\textwidth}
        \centering
        \includegraphics[width=0.98\textwidth]{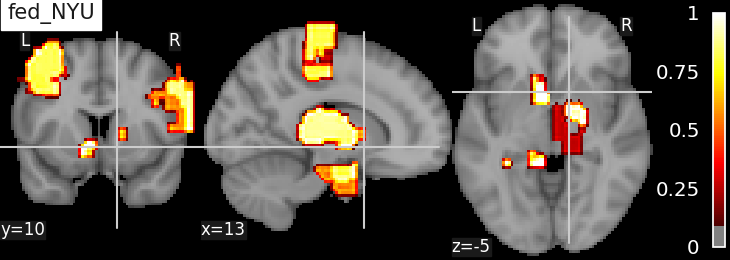}
        % \caption{Detected biomarkers for NYU site using \textbf{\textit{Fed}} strategy - view 1.}
    \end{subfigure}%
    ~
    \begin{subfigure}[t]{0.25\textwidth}
        \centering
        \includegraphics[width=0.98\textwidth]{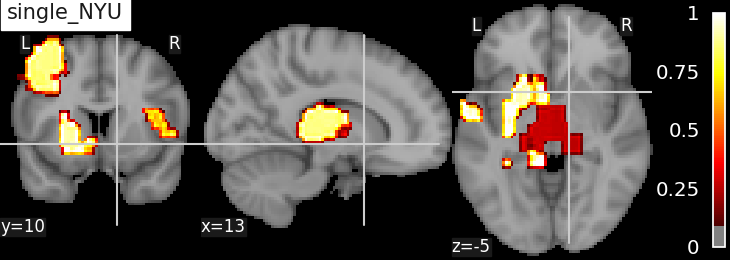}
        % \caption{Detected biomarkers for NYU site using \textbf{\textit{Fed}} strategy - view 2.}
    \end{subfigure}%
    ~
    \begin{subfigure}[t]{0.25\textwidth}
        \centering
        \includegraphics[width=0.98\textwidth]{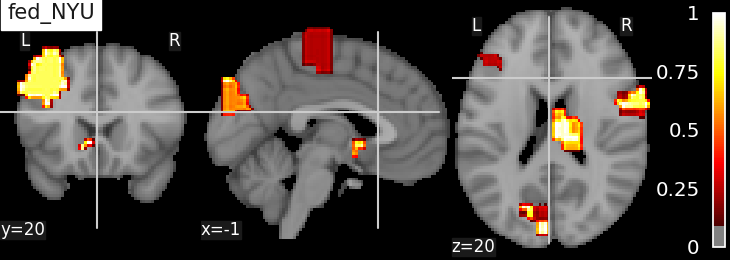}
        % \caption{Detected biomarkers for UM site using \textbf{\textit{Fed}} strategy - view 1.}
    \end{subfigure}%
    ~
    \begin{subfigure}[t]{0.25\textwidth}
        \centering
        \includegraphics[width=0.98\textwidth]{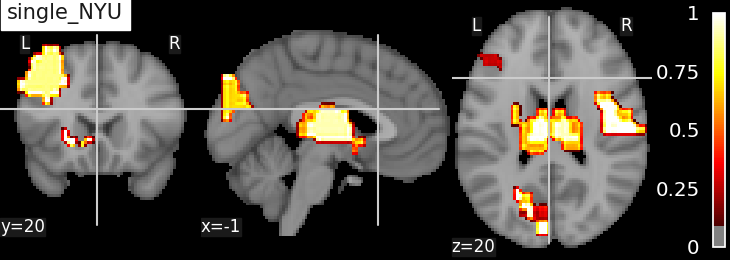}
        % \caption{Detected biomarkers for UM site using \textbf{\textit{Fed}} strategy - view 2.}
    \end{subfigure}%
    ~
    
    \begin{subfigure}[t]{0.25\textwidth}
        \centering
        \includegraphics[width=0.98\textwidth]{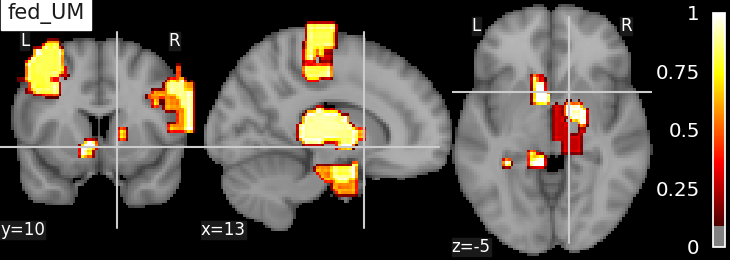}
        % \caption{Detected biomarkers for USM site using \textbf{\textit{Fed}} strategy - view 1.}
    \end{subfigure}%
    ~
    \begin{subfigure}[t]{0.25\textwidth}
        \centering
        \includegraphics[width=0.98\textwidth]{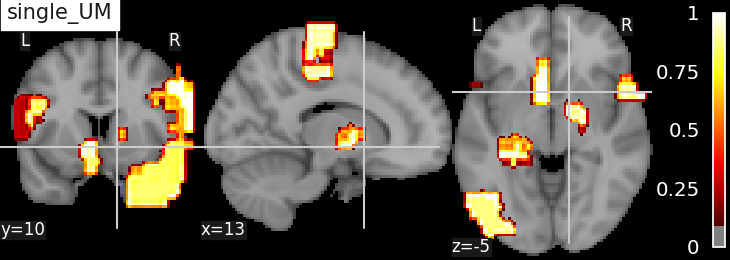}
        % \caption{Detected biomarkers for USM site using \textbf{\textit{Fed}} strategy - view 2.}
    \end{subfigure}%
    ~
    \begin{subfigure}[t]{0.25\textwidth}
        \centering
        \includegraphics[width=0.98\textwidth]{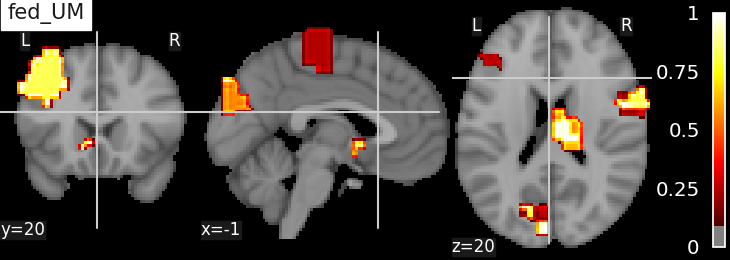}
        % \caption{Detected biomarkers for UCLA site using \textbf{\textit{Fed}} strategy - view 1.}
    \end{subfigure}%
    ~
    \begin{subfigure}[t]{0.25\textwidth}
        \centering
        \includegraphics[width=0.98\textwidth]{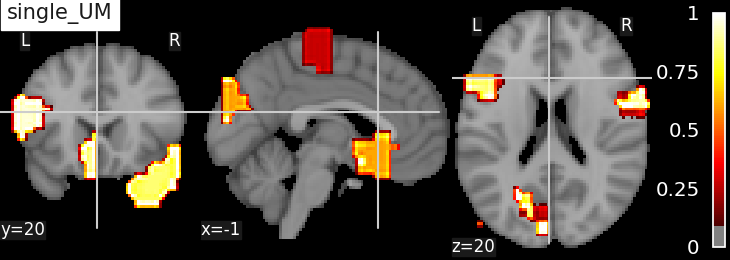}
        % \caption{Detected biomarkers for UCLA site using \textbf{\textit{Fed}} strategy - view 2.}
    \end{subfigure}%
    ~
    % \caption{Interpreting brain biomarkers associated with identifying ASD from federated learning model (\textit{Fed})} 
    % \label{fig:fedbio}
% \end{figure}

% \begin{figure}[t]
    % \centering
    \begin{subfigure}[t]{0.25\textwidth}
        \centering
        \includegraphics[width=0.98\textwidth]{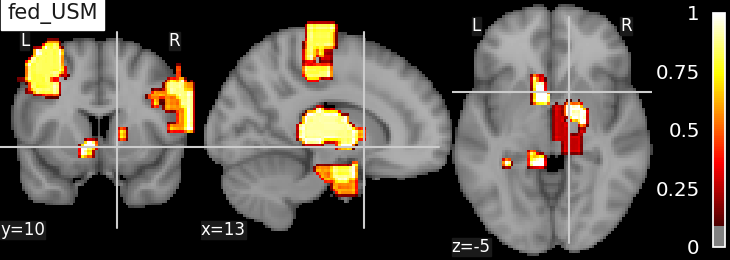}
        % \caption{Detected biomarkers for NYU site using \textbf{\textit{Single}} strategy - view 1.}
    \end{subfigure}%
    ~
    \begin{subfigure}[t]{0.25\textwidth}
        \centering
        \includegraphics[width=0.98\textwidth]{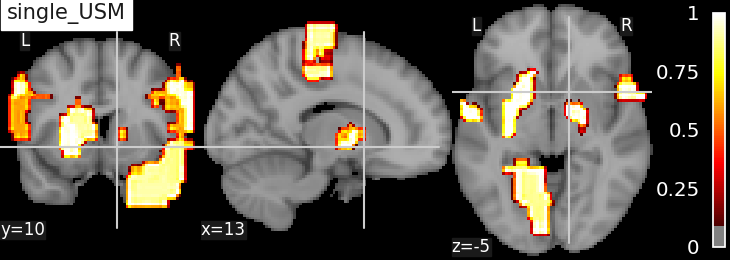}
        % \caption{Detected biomarkers for NYU site using \textbf{\textit{Single}} strategy - view 2.}
    \end{subfigure}%
    ~
    \begin{subfigure}[t]{0.25\textwidth}
        \centering
        \includegraphics[width=0.98\textwidth]{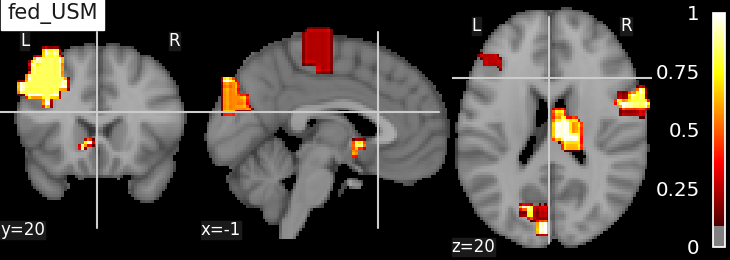}
        % \caption{Detected biomarkers for UM site using \textbf{\textit{Single}} strategy - view 1.}
    \end{subfigure}%
    ~
    \begin{subfigure}[t]{0.25\textwidth}
        \centering
        \includegraphics[width=0.98\textwidth]{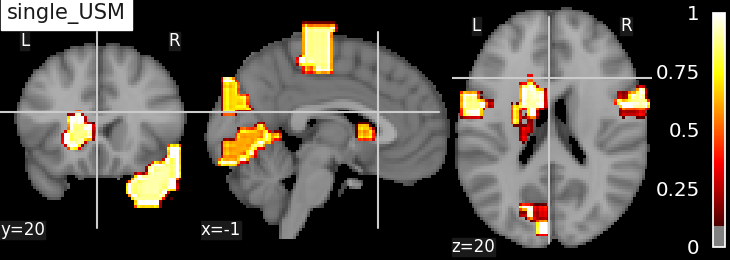}
        % \caption{Detected biomarkers for UM site using \textbf{\textit{Single}} strategy - view 2.}
    \end{subfigure}%
    ~
    
    \begin{subfigure}[t]{0.25\textwidth}
        \centering
        \includegraphics[width=0.98\textwidth]{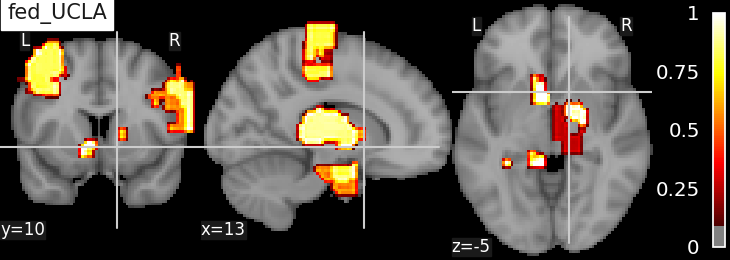}
        \caption{Biomarkers using \textbf{\textit{Fed}} strategy - view 1.}
    \end{subfigure}%
    ~
    \begin{subfigure}[t]{0.25\textwidth}
        \centering
        \includegraphics[width=0.98\textwidth]{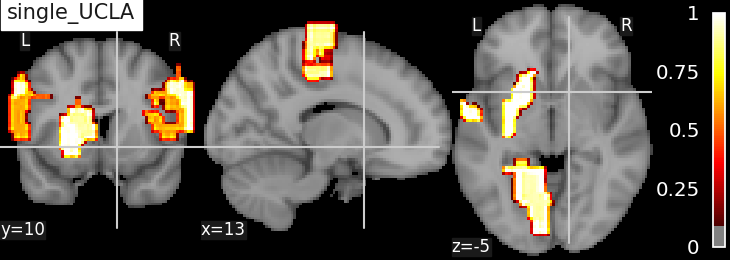}
        \caption{Biomarkers using \textbf{\textit{Single}} strategy - view 1.}
    \end{subfigure}%
    ~
    \begin{subfigure}[t]{0.25\textwidth}
        \centering
        \includegraphics[width=0.98\textwidth]{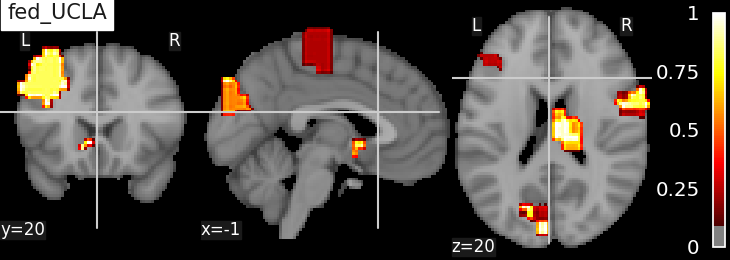}
        \caption{Biomarkers using \textbf{\textit{Fed}} strategy - view 2.}
    \end{subfigure}%
    ~
    \begin{subfigure}[t]{0.25\textwidth}
        \centering
        \includegraphics[width=0.98\textwidth]{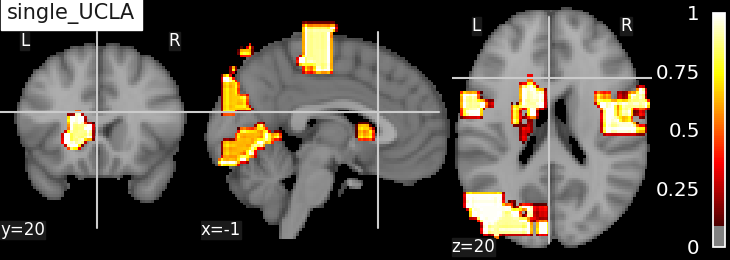}
        \caption{Biomarkers using \textbf{\textit{Single}} strategy - view 2.}
    \end{subfigure}%
    ~
    \caption{Interpreting brain biomarkers associated with identifying \textbf{ASD} from federated learning model (\textit{Fed}) and using single site data for training (\textit{Single}). {The colors stand for the relative importance scores of the ROIs and the values are denoted on the color bar. The names of the strategies and sites are denoted on the left-upper corners of each subfigure. Each row shows the results of NYU, UM, USM, UCLA site from top to bottom.}} 
    \label{fig:asdbio}
\end{figure*}

\begin{table}[]
\centering
\resizebox{0.5\textwidth}{!}{%
\begin{tabular}{@{}ll|cccccc@{}} \toprule
 &  & \textbf{Sementic} & \textbf{Comprehension} & \textbf{Social} & \textbf{Attention} & \textbf{Memory} & \textbf{Reward} \\  \hline
\multirow{3}{*}{Fed} & HC & 0.054 & 0.096 & 0.099 & 0.088 & 0.009 & -0.078 \\
 & ASD & -0.048 & -0.035 & -0.081 & 0.007 & 0.031 & 0.017 \\ 
&$\mathbf{|\Delta|}$ & \textbf{0.102}& \textbf{0.131}& \textbf{0.180}& \textbf{0.081}& \textbf{0.022}& \textbf{0.095}  \\ \hline
\multirow{3}{*}{Single} & HC & 0.050 & 0.043 & 0.069 & 0.053 & 0.022 & -0.062 \\
 & ASD & -0.029 & -0.005 & -0.094 & 0.005 & 0.041 & 0.010 \\ &$\mathbf{|\Delta|}$ &\textbf{0.079} & \textbf{0.048}& \textbf{0.163}& \textbf{0.048}& \textbf{0.019}& \textbf{0.072} \\ \bottomrule
\end{tabular}%
}
\small{$|\Delta|$ is the absolute difference between the scores of HC and ASC groups.}
\caption{Correlations between the detected biomarkers and functional keywords maps decoded by Neurosynth.}
\label{tab:decode}
\end{table}
Note that we defined "informativeness" as the difference of functional representations between ASD and HC groups and  "robustness" as the consistency of biomarker detection results across 4 sites. Whether informative and robust biomarkers can be interpreted is another dimension to evaluate a deep learning model apart from using accuracy-related metrics.  Here, we used the guided back-propagation method (Eq. \ref{eq:grad}) to interpret feature importance on \textit{Fed} and \textit{Single} model separately. The features of inputs were the functional connectivity between brain ROIs. First, we calculated $ g^c = [g_k^c]_{k=1}^{6105}$ for each testing point. To get the ROI level evaluation, we built a symmetric \textit{grad} matrix $\mathcal{G}$ where the $ij${th} entry is the $g^c$ of functional connectivity between ROI $i$ and $j$. We summed $\mathcal{G}$ over columns resulting in a 111-dim vector $s^c$ standing for the importance score of the 111 ROIs. We normalized $s^c=[s_r^c]_{r=1}^{111}$ by dividing  $ max(s^{c})$ to bound it to $[0,1]$. We averaged the results for all the test data points in each site. The ROIs with the top 10 important scores for classification (2 classes) and normalized importance scores on the ROIs were plotted for HC (Figure \ref{fig:hcbio}) and ASD (Figure \ref{fig:asdbio}). \textit{Fed} detected \textbf{robust} biomarkers across 4 sites, while the biomarkers detected by \textit{Single} were different across different sites. Further, we listed the correlations between the biomarkers with functional keyword maps in Table \ref{tab:decode} by Neurosynth \citep{yarkoni2011large}. The biomarkers detected by \textit{Fed} were more distinguishable than those of \textit{Single}, as the differences between correlation values for HC and ASD group were larger than those of \textit{Single} (see the $|\Delta|$ scores of \textit{Fed} are larger than those of \textit{Single} in Table \ref{tab:decode}). Therefore, we found the biomarkers detected by \textit{Fed} were more informative. We could infer from Table \ref{tab:decode} that the semantic, comprehension, social and attention-related functional connectivity was more salient in the HC group, while memory and reward-related functional connectivity was more salient in the ASD group. Hence, the biomarkers detected by \textit{Fed} were more robust and informative. The names of the biomarkers of each group detected by \textit{Fed} and \textit{Single} were listed in the Appendix. 

\subsection{Limitation and discussion}
Although, based on our empirical investigation that the communication pace, which controled how often the local and global model update the weight information, did not affect the classification performance, we could not draw the conclusion that the pace parameter was irrelevant. A more extensive range of pace values should be examined according to the application. Also, we used practical approaches to investigate privacy-preserving mechanisms. However, the sensitivity of the mapping function $h : \mathcal{D} \rightarrow \mathbb{R}^m$, the deep learning classifier in our case, was difficult to estimate. Hence, we did not explicitly give the bound $\epsilon$. A recent study \citep{zhu2019deep} also demonstrated Gaussian and Laplace noise higher than a certain scale can be a good defense to reconstruction attack. According to the specific application and dataset, we can empirically estimate a suitable noise level from attacking perspective as well. In our experiments, we used the averaging method to incorporate models' outputs for \textit{Ensemble}. To achieve better performance for \textit{Ensemble}, more advanced ensemble methods could be exploited, such as gradient tree boosting, stacking, and forest of randomized trees \citep{zhou2012ensemble}.  We evaluated the biomarkers at the ROI-level. The functional connectivity also could be used as biomarkers. More advanced deep learning models can be explored as well. In order to show the strong direct associations between the biomarkers and disease diagnosis or treatment outcome prediction, down-stream tasks such as regression to ADOS scores using the biomarkers are worthy of exploring. We found that domain adaptation methods were not always a beneficial addition to the federated model. Based on the results presented in Section \ref{sec:startegyres} and Section \ref{sec:interpres}, we found that domain adaption techniques improved the classification accuracy of some sites but not all (two out of four sites were better and one site kept the same for $MoE$ and $MoE$ by comparing mean accuracies). This could be because the current model updating strategy is not optimal.  Going forward, we plan to examine the distribution of the latent features of different data owners first, then decide whether to adopt our proposed domain adaptation methods. In contrast to other FL applications, such as typing recommendations in Google and Apple, where there might be millions of FL participants, there are many fewer participants in multi-site medical data analysis. Hence, the number of FL participants might play a role here, especially if we use an averaging strategy to update the global model. Incorporating more advanced model-selection and updating strategies will help avoid including the wrong private model in updating \citep{mohri2019agnostic,nishio2019client}.  

\section{Conclusion} \label{sec:conclusion}
{In this work, we have presented a privacy-preserving federated learning framework for multi-site fMRI analysis. We have investigated the communication pace and the privacy-preserving randomized mechanisms for the problem of using brain functional connectivity to classify ASD and HC. To overcome the domain shift issue, we have proposed two strategies: MoE and adversarial domain alignment to boost federated learning model performance. We have also evaluated the deep learning model for neuroimaging from the biomarker detection perspective.

Our results have demonstrated the advantage of using a federated framework to utilize multi-site data without data sharing compared to alternative methods. We have shown federated learning performance can potentially be boosted by adding domain adaptation and discussed the condition of benefits. In addition, the proposed federated learning model has revealed possible brain biomarkers for identifying ASD. Our work also has broader implications into other disease areas, particularly rare diseases with fewer patients. In these situations, utilizing data across multiple sites is critical and required for meaningful conclusions.

Our approach brings new hope for accelerating deep learning applications in the field of medical imaging, where data isolation and the emphasis on data privacy have become challenges. It can establish a unified model for multiple medical institutions while protecting local data, allowing medical institutions to work together with the required data security.}

%It is expected that in the near future, federated learning would break the barriers between institutions and establish a community where data and knowledge could be shared together with safety, and the benefits would be fairly distributed according to the contribution of eachparticipant. The bonus of artificial intelligence would finally be brought to every corner of our lives.

\section{Declaration of Competing Interest}
The authors declare that they have no known competing financial
interests or personal relationships that could have appeared to
influence the work reported in this paper.

\section{Acknowledgements}
Data collection and sharing for this project was funded by the
Autism Brain Imaging Data Exchange dataset (ABIDE) \citep{di2014autism}. 
Parts of this research was supported by National Institutes of Health (NIH) [R01NS035193, R01MH100028].

\bibliographystyle{model2-names.bst}\biboptions{authoryear}
\bibliography{ref.bib}

\clearpage
\newpage
\section*{Appendix}
\subsection*{\textbf{Architecture of the models}}
We provide the detailed model architecture for each strategy we used in our study. For each fully connected (FC), we provide the input and output dimension. For drop-out (Dropout) layers, we provide the probability of an element to be zeroed. We denote batch normalization layers as (BN), relu layers as (ReLU) and softmax layers as Softmax.\\

Models for \textit{Single}, \textit{Cross} and \textit{Ensemble} are shown in Table \ref{tab:mlp1}.
\begin{table}[htpb]
\centering
\resizebox{0.4\textwidth}{!}{%
\begin{tabular}{l|l} \toprule
\multicolumn{1}{c|}{\textbf{Layer}} & \multicolumn{1}{c}{\textbf{Configuration}} \\ \midrule
\multicolumn{2}{c}{MLPs} \\ \hline
1 & Dropout (0.5), FC (6105, 8), ReLU, BN \\ \hline
2 & Dropout (0.5), FC (8, 2), Softmax \\ \bottomrule
\end{tabular}%
}
\caption{Model architecture for ABIDE rs-fMRI classification task under \textit{Single}, \textit{Cross} and \textit{Ensemble} strategies.}
\label{tab:mlp1}
\end{table}

Models for \textit{Cross} and \textit{Ensemble} is shown in Table \ref{tab:mlp2}.
\begin{table}[htpb]
\centering
\resizebox{0.4\textwidth}{!}{%
\begin{tabular}{l|l} \toprule
\multicolumn{1}{c|}{\textbf{Layer}} & \multicolumn{1}{c}{\textbf{Configuration}} \\ \midrule
\multicolumn{2}{c}{MLPs} \\ \hline
1 & Dropout (0.5), FC (6105, 16), ReLU, BN \\ \hline
2 & Dropout (0.5), FC (16, 2),  Softmax \\ \bottomrule
\end{tabular}%
}
\caption{Model architecture for ABIDE rs-fMRI classification task under \textit{Fed} and \textit{Mix} strategies.}
\label{tab:mlp2}
\end{table}
    
Models for \textit{Fed-MoE} strategy is shown in Table \ref{tab:moe}.
\begin{table}[htpb]
\centering
\resizebox{0.4\textwidth}{!}{%
\begin{tabular}{l|c|l|c} \toprule
\multicolumn{1}{c|}{\textbf{Layer}} & \textbf{Configuration} & \textbf{Layer} & \textbf{Configuration} \\ \hline
\multicolumn{2}{c|}{Private Model} & \multicolumn{2}{c}{Global Model} \\ \hline
1 & \begin{tabular}[c]{@{}c@{}}Dropout (0.5), \\ FC (6105,8),\\  ReLU, BN\end{tabular} & 1 & \begin{tabular}[c]{@{}c@{}}Dropout (0.5), \\ FC (6105,16), \\ ReLU, BN\end{tabular} \\ \hline
2 & \begin{tabular}[c]{@{}c@{}}Dropout (0.5), \\ FC (8,2)\end{tabular} & 2 & \begin{tabular}[c]{@{}c@{}}Dropout (0.5),\\  FC (16,2) \end{tabular} \\ \midrule
\textbf{Layer} & \multicolumn{3}{c}{\textbf{Configuration}} \\ \hline
\multicolumn{4}{c}{MoE} \\ \hline
1 & \multicolumn{3}{c}{FC (2,1), Sigmoid} \\ \bottomrule
\end{tabular}%
}
\caption{Model architecture for ABIDE rs-fMRI classification task under Fed-MoE strategy.}
\label{tab:moe}
\end{table}

Models for \textit{Fed-Align} strategy is shown in Table \ref{tab:align}
\begin{table}[htpb]
\centering
\resizebox{0.4\textwidth}{!}{%
\begin{tabular}{l|c} \toprule
\textbf{Layer} & \textbf{Configuration} \\ \hline
\multicolumn{2}{c}{Feature Generator} \\ \hline
1 & Dropout (0.5), FC (6105, 16), ReLU, BN \\ \hline
\multicolumn{2}{c}{\textbf{Domain Discriminator}} \\ \hline
1 & FC (6105, 8), ReLU \\ \hline
2 & FC (8, 1), sigmoid \\ \hline
\multicolumn{2}{c}{\textbf{Classifier}} \\ \hline
1 & Dropout (0.5), FC (16, 2), Softmax \\ \bottomrule
\end{tabular}%
}
\caption{Model architecture for ABIDE rs-fMRI classification task under Fed-Align strategy.}
\label{tab:align}
\end{table}

\subsection*{\textbf{Names of the biomarkers}}
We list the top 10 important ROIs (plotted in Figure \ref{fig:hcbio} and Figure \ref{fig:asdbio} in descending order.\\

1. HC biomarkers detected by \textit{Fed}:
\begin{itemize}
    \item[NYU]: 'Right Heschl's Gyrus (includes H1 and H2)'
 'Right Inferior Temporal Gyrus' 'Right Superior Frontal Gyrus'
 'Right Precentral Gyrus' 'Left Intracalcarine Cortex'
 'Left Cingulate Gyrus' 'Left Temporal Pole'
 'Right Superior Temporal Gyrus' 'Right Middle Temporal Gyrus'
 'Left Planum Polare'
 \item[UM]: 'Right Heschl's Gyrus (includes H1 and H2)'
 'Right Inferior Temporal Gyrus' 'Right Superior Frontal Gyrus'
 'Right Precentral Gyrus' 'Left Intracalcarine Cortex'
 'Left Cingulate Gyrus' 'Right Superior Temporal Gyrus'
 'Left Temporal Pole' 'Right Middle Temporal Gyrus' 'Left Planum Polare'
 \item[USM]: 'Right Heschl's Gyrus (includes H1 and H2)'
 'Right Inferior Temporal Gyrus' 'Right Superior Frontal Gyrus'
 'Right Precentral Gyrus' 'Left Intracalcarine Cortex'
 'Left Cingulate Gyrus' 'Right Middle Temporal Gyrus' 'Left Planum Polare'
 'Right Superior Temporal Gyrus' 'Left Temporal Pole'
 \item[UCLA]: 'Right Heschl's Gyrus (includes H1 and H2)'
 'Right Superior Frontal Gyrus' 'Right Inferior Temporal Gyrus'
 'Right Precentral Gyrus' 'Left Intracalcarine Cortex'
 'Left Cingulate Gyrus' 'Left Temporal Pole'
 'Right Superior Temporal Gyrus' 'Right Middle Temporal Gyrus'
 'Left Planum Polare'
\end{itemize}{}

2. HC biomarkers detected by \textit{Single}:
\begin{itemize}
    \item[NYU]: 'Right Middle Temporal Gyrus' 'Right Occipital Pole'
 'Right Supramarginal Gyrus' 'Left Paracingulate Gyrus'
 'Right Precentral Gyrus' 'Right Frontal Orbital Cortex'
 'Right Temporal Pole' 'Right Frontal Medial Cortex'
 'Right Parahippocampal Gyrus' 'Left Parietal Operculum Cortex'
 \item[UM]: 'Right Supramarginal Gyrus' 'Right Middle Temporal Gyrus'
 'Left Inferior Temporal Gyrus' 'Left Inferior Frontal Gyrus'
 'Left Paracingulate Gyrus' "Right Heschl's Gyrus (includes H1 and H2)"
 'Left Lingual Gyrus' 'Right Superior Temporal Gyrus'
 'Left Frontal Orbital Cortex' 'Left Superior Temporal Gyrus'
 \item[USM]: 'Right Middle Temporal Gyrus' 'Right Supramarginal Gyrus'
 'Left Superior Temporal Gyrus' 'Right Occipital Pole'
 'Left Paracingulate Gyrus' 'Right Precentral Gyrus'
 'Left Inferior Temporal Gyrus' 'Left Middle Temporal Gyrus' 'None'
 'Left Hippocampus'
 \item[UCLA]: 'Right Supramarginal Gyrus' 'Right Middle Temporal Gyrus'
 'Right Occipital Pole' 'Right Precentral Gyrus'
 'Right Temporal Occipital Fusiform Cortex' 'Left Lingual Gyrus'
 'Left Inferior Temporal Gyrus' 'Left Paracingulate Gyrus'
 'Right Cingulate Gyrus' 'Left Inferior Frontal Gyrus'
\end{itemize}{}

3. ASD biomarkers detected by \textit{Fed}:
\begin{itemize}
    \item[NYU]: 'Left Accumbens' 'Left Parahippocampal Gyrus' 'Right Thalamus'
 "Right Heschl's Gyrus (includes H1 and H2)" 'Right Pallidum'
 'Left Middle Frontal Gyrus' 'Right Precentral Gyrus'
 'Right Parahippocampal Gyrus' 'Left Cuneal Cortex'
 'Left Temporal Fusiform Cortex'
 \item[UM]: 'Left Accumbens' 'Left Frontal Operculum Cortex' 'Right Thalamus'
 'Right Lateral Occipital Cortex' 'Right Pallidum'
 'Left Postcentral Gyrus'
 'Left Juxtapositional Lobule Cortex (formerly Supplementary Motor Cortex)'
 'Right Middle Frontal Gyrus' 'Right Occipital Fusiform Gyrus'
 'Left Central Opercular Cortex'
 \item[USM]: 'Left Accumbens' 'Left Frontal Operculum Cortex' 'Right Thalamus'
 'Right Pallidum' 'Right Lateral Occipital Cortex'
 'Left Juxtapositional Lobule Cortex (formerly Supplementary Motor Cortex)'
 'Left Postcentral Gyrus' 'Right Middle Frontal Gyrus'
 'Left Central Opercular Cortex' 'Right Occipital Fusiform Gyrus'
 \item[UCLA]: 'Left Accumbens' 'Left Frontal Operculum Cortex' 'Right Thalamus'
 'Right Lateral Occipital Cortex' 'Right Pallidum'
 'Left Juxtapositional Lobule Cortex (formerly Supplementary Motor Cortex)'
 'Left Postcentral Gyrus' 'Right Middle Frontal Gyrus'
 'Right Occipital Fusiform Gyrus' 'Left Central Opercular Cortex'
\end{itemize}{}

4. ASD biomarkers detected by \textit{Single}:
\begin{itemize}
    \item[NYU]: 'Right Occipital Fusiform Gyrus' 'Left Angular Gyrus' 'Left Putamen'
 'Left Thalamus' 'Left Supracalcarine Cortex' 'Right Cingulate Gyrus'
 'Left Frontal Operculum Cortex'
 'Left Juxtapositional Lobule Cortex (formerly Supplementary Motor Cortex)'
 'Right Thalamus' 'Left Accumbens'
 \item[UM]: 'Left Supracalcarine Cortex' 'Left Accumbens'
 'Right Middle Frontal Gyrus' 'Left Temporal Fusiform Cortex'
 'Right Occipital Fusiform Gyrus' 'Left Parahippocampal Gyrus'
 'Left Subcallosal Cortex' 'Left Hippocampus' 'Left Middle Frontal Gyrus'
 'Right Pallidum'
 \item[USM]: 'Right Occipital Fusiform Gyrus' 'Left Putamen' 'Right Cingulate Gyrus'
 'Right Middle Frontal Gyrus' 'Left Middle Frontal Gyrus'
 'Right Temporal Pole' 'Left Caudate' 'Right Pallidum'
 'Right Central Opercular Cortex' 'Right Paracingulate Gyrus'
 \item[UCLA]: 'Left Putamen' 'Right Occipital Fusiform Gyrus' 'Right Cingulate Gyrus'
 'Right Temporal Pole' 'Right Central Opercular Cortex' 'Left Caudate'
 'Right Paracingulate Gyrus' 'Right Parahippocampal Gyrus'
 'Right Middle Frontal Gyrus' 'Left Angular Gyrus'
\end{itemize}{}

\subsection*{\textbf{Additional experiments using alternative atlas}} 
Following the preprocessing pipeline in Section 4.1.2, we replaced the structural Harvard-Oxford (HO) atlas with a functional atlas, the Craddock 200 (CC200), which parcellates brain into 200 ROIs. We replicated the main experiment of our work as shown in Section 4.3, but changed the input dimension of each MLP from 6105 to $19900(=200\times100-100)$. The comparison results were shown in Appendix Table \ref{tab:strategy}. Overall, the accuracies of using CC200 was lower than the accuracies of using HO. This could be caused by overfitting as the number of parameters required for training CC200 data is higher than HO due to the dimensionality of the input. In \textit{Cross}, we denoted the site used for training as 'tr$<$site$>$'. As the testing data were all the other whole sites, there was no standard deviation (std) to report. Also, we ignored the performance of the site used for training. The other results were reported using the 'mean (std)' format. By comparing the mean accuracy only, we highlighted the best accuracy in Table \ref{tab:strategy}. \textit{Fed-MoE} outperformed \textit{Fed} on NYU, UM and UCLA site and \textit{Fed-Align} outperformed \textit{Fed} on NYU, UM and UCLA site by comparing mean accuracies. All \textit{Fed} and \textit{Fed+Domain Adaptation} strategies showed significant improvement compared to \textit{Cross}, \textit{Single} and \textit{Ensemble} strategies. The results showed the replicability of our method on a different atlas.
        \begin{table}[htpb]
            \centering
            \resizebox{0.45\textwidth}{!}{%
            \begin{tabular}{@{}l|cccc@{}} \toprule
                   & \textbf{NYU} & \textbf{UM}  & \textbf{USM} & \textbf{UCLA} \\  \hline
            trNYU  & -            & 0.559        & 0.669       & 0.689         \\
            trUM   & 0.611        & -            & 0.577        & 0.492         \\
            trUSM  & 0.635        & 0.693        & -            & 0.587         \\
            trUCLA & 0.580        & 0.681        & 0.750        & -             \\
            Single & 0.610(0.062) & 0.681(0.071) & 0.689(0.083) & 0.536(0.121)  \\
            Ensemble &0.626(0.054)& 0.676(0.053)  & 0.694(0.092)  &  0.649(0.069) \\
            Fed    & 0.648(0.157) & 0.693(0.055) & \textbf{0.789(0.153)} & 0.652(0.098)  \\
            Fed-MoE    & 0.664(0.072) & \textbf{0.705(0.070)} & 0.756(0.061) & 0.663(0.146) \\
            Fed-Align  &\textbf{0.683(0.097)} & 0.694(0.109) & 0.712(0.053) & 0.667(0.106)  \\
            Mix    & 0.652(0.052) & 0.690(0.092) & 0.752(0.090) & \textbf{0.725(0.130)}  \\ \bottomrule
            \end{tabular}%
            }
            \caption{Results of using different training strategies with atlas CC200. Numbers shown are  mean classification accuracies and corresponding stds.}
            \label{tab:strategy}
            \end{table}

\subsection*{\textbf{Additional experiments on ASD sex classification}}
Following the preprocessing pipeline in Section 4.1.2, in an effort to illustrate that our approach works effectively on another problem, we present new results on gender classification for ASD subjects. We replicated the main experiment of our work as shown in Section 4.3, but excluded USM site as it only contained male ASDs. The comparison results were shown in Table \ref{tab:strategy2}. Strategy \textit{Basline} was denoted as the random guess accuracy. We found obvious improvement in using federated related strategies and \textit{Mix} strategy, whereas the other strategies that only could train the model using the data in a single site did not show improvement.
        \begin{table}[htpb]
            \centering
            \resizebox{0.35\textwidth}{!}{%
            \begin{tabular}{@{}l|ccc@{}} \toprule
                   & \textbf{NYU} & \textbf{UM}  &  \textbf{UCLA} \\  \hline
            trNYU  & -            & 0.837        & 0.838             \\
            trUM   & 0.877       & -            & 0.865         \\
            trUCLA & 0.890        & 0.814            & -             \\
            Single &0.879(0.010)& 0.902(0.013)    &  0.887(0.048) \\
            Ensemble & 0.915(0.029) & 0.917(0.048) & 0.891(0.062)  \\
            Fed    & 0.972(0.033) & 0.913(0.054) &  0.925(0.098)  \\
            Fed-MoE    & 0.967(0.033)  & 0.917(0.048) & 0.937(0.108) \\
            Fed-Align   & 0.973(0.028) & 0.906(0.048) & 0.955(0.054)  \\
            Mix    & 0.983(0.029) & 0.925(0.052) & 0.975(0.059)  \\ 
            Baseline & 0.857 &0.890 &0.838 \\ \bottomrule
            \end{tabular}%
            }
            \caption{Results of ASD sex classification using different training strategies. Numbers shown are mean classification accuracies and corresponding stds.}
            \label{tab:strategy2}
            \end{table}
\begin{figure*}[htpb]
    \centering
    \begin{subfigure}[t]{0.25\textwidth}
        \centering
        \includegraphics[width=0.98\textwidth]{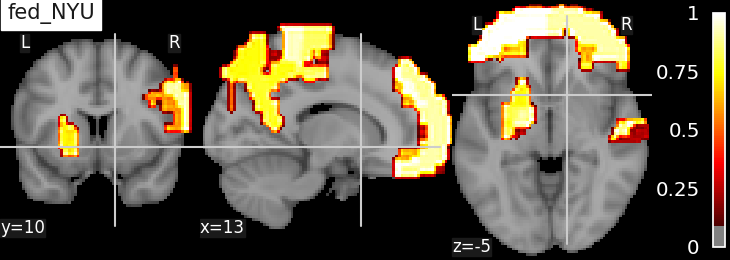}
        % \caption{Detected biomarkers for NYU site using \textbf{\textit{Fed}} strategy - view 1.}
    \end{subfigure}%
    ~
    \begin{subfigure}[t]{0.25\textwidth}
        \centering
        \includegraphics[width=0.98\textwidth]{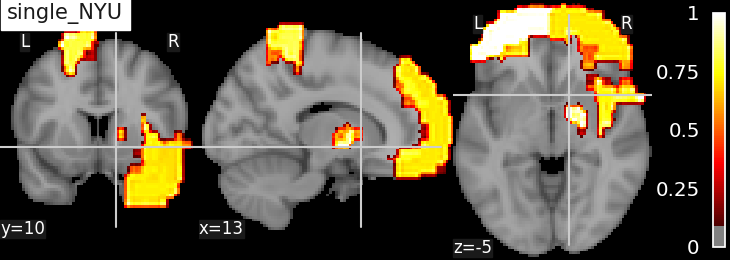}
        % \caption{Detected biomarkers for NYU site using \textbf{\textit{Fed}} strategy - view 2.}
    \end{subfigure}%
    ~
    \begin{subfigure}[t]{0.25\textwidth}
        \centering
        \includegraphics[width=0.98\textwidth]{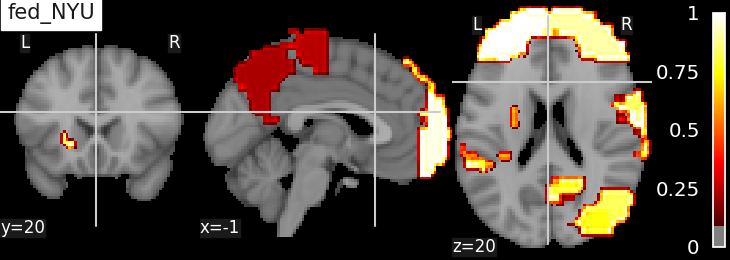}
        % \caption{Detected biomarkers for UM site using \textbf{\textit{Fed}} strategy - view 1.}
    \end{subfigure}%
    ~
    \begin{subfigure}[t]{0.25\textwidth}
        \centering
        \includegraphics[width=0.98\textwidth]{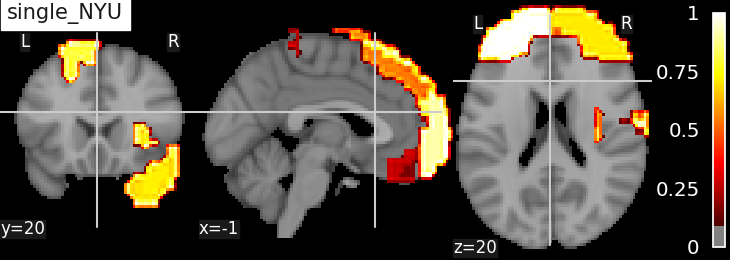}
        % \caption{Detected biomarkers for UM site using \textbf{\textit{Fed}} strategy - view 2.}
    \end{subfigure}%
    ~
    
    \begin{subfigure}[t]{0.25\textwidth}
        \centering
        \includegraphics[width=0.98\textwidth]{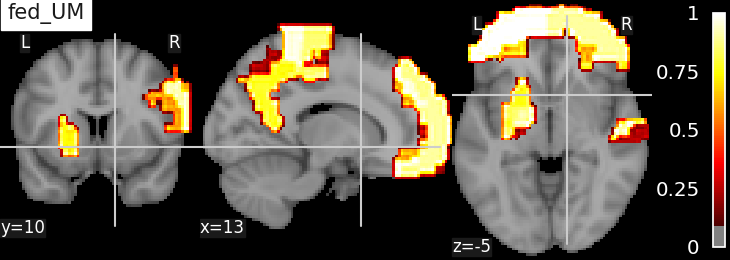}
        % \caption{Detected biomarkers for USM site using \textbf{\textit{Fed}} strategy - view 1.}
    \end{subfigure}%
    ~
    \begin{subfigure}[t]{0.25\textwidth}
        \centering
        \includegraphics[width=0.98\textwidth]{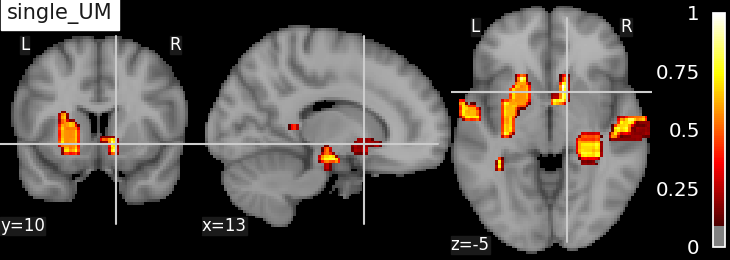}
        % \caption{Detected biomarkers for USM site using \textbf{\textit{Fed}} strategy - view 2.}
    \end{subfigure}%
    ~
    \begin{subfigure}[t]{0.25\textwidth}
        \centering
        \includegraphics[width=0.98\textwidth]{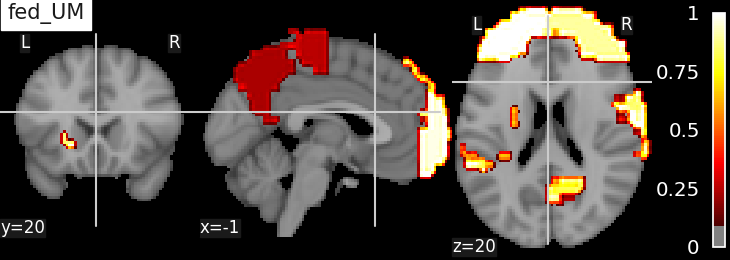}
        % \caption{Detected biomarkers for UCLA site using \textbf{\textit{Fed}} strategy - view 1.}
    \end{subfigure}%
    ~
    \begin{subfigure}[t]{0.25\textwidth}
        \centering
        \includegraphics[width=0.98\textwidth]{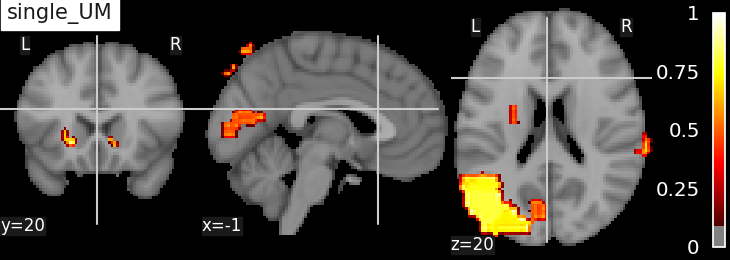}
        % \caption{Detected biomarkers for UCLA site using \textbf{\textit{Fed}} strategy - view 2.}
    \end{subfigure}%
    ~
    
    \begin{subfigure}[t]{0.25\textwidth}
        \centering
        \includegraphics[width=0.98\textwidth]{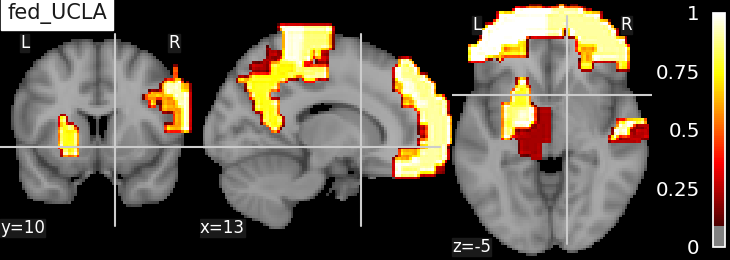}
        \caption{Biomarkers using \textbf{\textit{Fed}} strategy - view 1.}
    \end{subfigure}%
    ~
    \begin{subfigure}[t]{0.25\textwidth}
        \centering
        \includegraphics[width=0.98\textwidth]{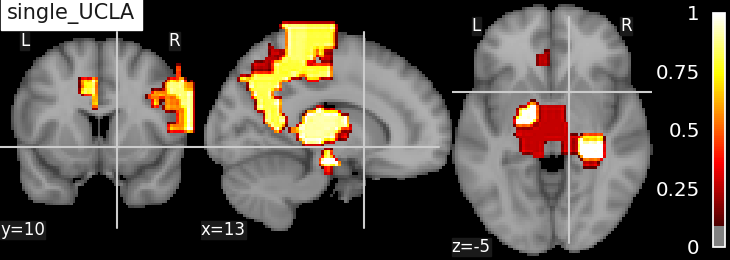}
        \caption{Biomarkers using \textbf{\textit{Single}} strategy - view 1.}
    \end{subfigure}%
    ~
    \begin{subfigure}[t]{0.25\textwidth}
        \centering
        \includegraphics[width=0.98\textwidth]{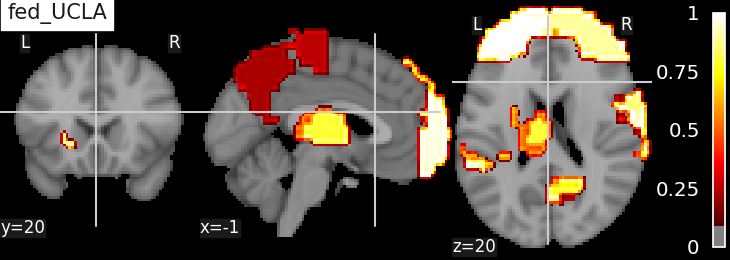}
        \caption{Biomarkers using \textbf{\textit{Fed}} strategy - view 2.}
    \end{subfigure}%
    ~
    \begin{subfigure}[t]{0.25\textwidth}
        \centering
        \includegraphics[width=0.98\textwidth]{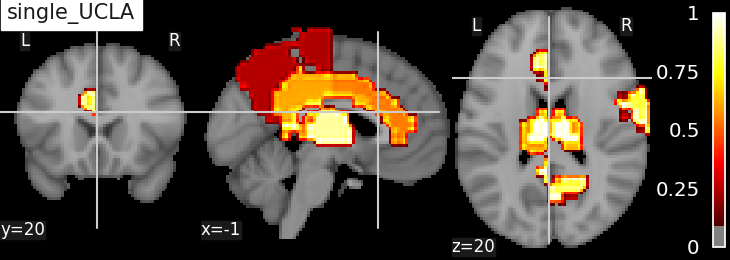}
        \caption{Biomarkers using \textbf{\textit{Single}} strategy - view 2.}
    \end{subfigure}%
    ~
    \caption{Interpreting brain biomarkers associated with identifying \textbf{Male} among ASD subjects from federated learning model (\textit{Fed}) and using single site data for training (\textit{Single}). The colors stand for the relative importance scores of the ROIs and the values are denoted on the color bar. The names of the strategies and sites are denoted on the left-upper corners of each subfigure. Each row shows the results of NYU, UM,  UCLA site from top to bottom.} 
    \label{fig:sexbio}
\end{figure*}

Following Section 4.4.3, we used gradient-based feature importance analysis method (Section 3.3) to calculate $ g^c = [g_k^c]_{k=1}^{6105}$ for each testing point firstly. To get the ROI level evaluation, we built a symmetric \textit{grad} matrix $\mathcal{G}$ where the $ij${th} entry was the $g^c$ of functional connectivity between ROI $i$ and $j$. We summed $\mathcal{G}$ over columns resulting in a 111-dim vector $s^c$ standing for the importance score of the 111 ROIs. We normalized $s^c=[s_r^c]_{r=1}^{111}$ by dividing  $ max(s^{c})$ to bound it to $[0,1]$. We averaged the results for all the test data points in each site.  The ROIs with the top 10 important scores for ASD subjects sex classification and normalized importance scores on the ROIs were plotted for male ASD (Figure \ref{fig:sexbio}).  The sex effects on Frontal Gyrus and Angular Gyrus were pointed out in recent work \citep{alaerts2016sex} and our methods highlighted those ROIs in Figure \ref{fig:sexbio}.
"
%\subsection*{Learning Curve}
\end{document}